\documentclass[preprint,12pt]{elsarticle}

\usepackage{fancyhdr}
\usepackage{algorithmic}
\usepackage{algorithm}
\usepackage{array}
\usepackage{natbib}
\usepackage[caption=false,font=small,labelfont=sf,textfont=sf]{subfig}
\usepackage{textcomp}
\usepackage{stfloats}
\usepackage{url}
\usepackage{verbatim}
\usepackage{graphicx}
\usepackage[usenames,dvipsnames]{color} %using the color package, not xcolor
\usepackage[export]{adjustbox}
\usepackage{amsmath,amsfonts}
\usepackage{caption}
\usepackage{subfig}    
\usepackage{natbib}
\setcitestyle{numbers,open={[},close={]}} %Citation-related commands
\usepackage{xcolor, soul}
\sethlcolor{green}
\usepackage[dvipsnames,svgnames,x11names]{xcolor}
\usepackage[marginparwidth=2.5cm]{geometry}
\usepackage[commandnameprefix=always, final]{changes}
 \definechangesauthor[color=blue]{AE}

\usepackage{lineno}
\begin{document}

\begin{frontmatter}

%% Title, authors and addresses

%% use the tnoteref command within \title for footnotes;
%% use the tnotetext command for theassociated footnote;
%% use the fnref command within \author or \affiliation for footnotes;
%% use the fntext command for theassociated footnote;
%% use the corref command within \author for corresponding author footnotes;
%% use the cortext command for theassociated footnote;
%% use the ead command for the email address,
%% and the form \ead[url] for the home page:
%% \title{Title\tnoteref{label1}}
%% \tnotetext[label1]{}
%% \author{Name\corref{cor1}\fnref{label2}}
%% \ead{email address}
%% \ead[url]{home page}
%% \fntext[label2]{}
%% \cortext[cor1]{}
%% \affiliation{organization={},
%%             addressline={},
%%             city={},
%%             postcode={},
%%             state={},
%%             country={}}
%% \fntext[label3]{}

\title{Multi-Modal World Model for Physical Robot Interactions: Simultaneous Visual and Tactile Predictions for Enhanced Accuracy}
%Simultaneous Visual and Tactile Predictions Boost Accuracy}
%Boosted Prediction Accuracy through Integration of Visual and Tactile Modalities}

%% use optional labels to link authors explicitly to addresses:
%% \author[label1,label2]{}
%% \affiliation[label1]{organization={},
%%             addressline={},
%%             city={},
%%             postcode={},
%%             state={},
%%             country={}}
%%
%% \affiliation[label2]{organization={},
%%             addressline={},
%%             city={},
%%             postcode={},
%%             state={},
%%             country={}}

%\thanks{$^1$ School of Computer Science, University of Lincoln, UK}
%\thanks{$^2$ University of Surrey, UK}
%\thanks{Corresponding author: a.ghalamzan@surrey.ac.uk}}

\author[label1]{Willow Mandil} %% Author name
\affiliation[label1]{organization={University of Lincoln},
            addressline={Brayford Pool},             city={Lincoln},
             postcode={LN6 7DQ},
             %state={},             
             country={UK}}
%% Author affiliation
\author[label2,label3]{Amir Ghalamzan E.} %% Author name
\affiliation[label2]{organization={University of Sheffield},
            addressline={Mapin Building},             city={Sheffield},
             postcode={S1 3JD},
             %state={}             
             country={UK, }}
\affiliation[label3]{Corresponding Author}
%% Abstract
\begin{abstract}
%% Text of abstract
Predicting the outcomes of robotic actions, often referred to as learning a \emph{world model}, in complex environments remains a fundamental challenge in robotics. Existing approaches primarily rely on visual observations and action inputs to generate video-based predictions, frequently overlooking the critical role of tactile feedback in understanding physical interactions. In this work, we investigate the integration of tactile and visual information within predictive perception systems for physical robot interaction. We demonstrate that visuo–tactile prediction provides the greatest benefits in physically ambiguous interaction regimes, while improvements are naturally limited when object dynamics are visually inferable. Furthermore, we introduce two novel robot-pushing datasets collected using a magnetic-based tactile sensor for unsupervised learning. The first dataset comprises visually identical objects with varying physical properties, explicitly isolating physical ambiguity, while the second mirrors existing robot-pushing benchmarks involving clusters of household objects. Our results show that tactile–visual integration improves prediction accuracy and robustness under physical ambiguity, while offering limited gains in visually unambiguous settings. Code and datasets are publicly available\footnote{{\tt https://sites.google.com/view/spots-research} and {\tt https://github.com/imanlab/WM-4-PRI}}. 
\end{abstract}

%%Graphical abstract
%\begin{graphicalabstract}
%\includegraphics{grabs}
%\end{graphicalabstract}

%%Research highlights
%\begin{highlights}
%\item Research highlight 1
%\item Research highlight 2
%\end{highlights}

%% Keywords
\begin{keyword}
World Model \sep Deep Learning in Robotics \sep Perception for Manipulation \sep Model Learning for Control

\end{keyword}

\end{frontmatter}

%%%%%%%%%%%%%%%%%%%%%%%%%%%%%%%%%%%%%%%%%
\section{Introduction}

Physical interaction is a fundamental aspect of human experience, shaping our ability to navigate and manipulate the world around us. As robots are increasingly deployed in real-world environments with more complex tasks, physical robot interactions (\textbf{PRI}) will become a critical feature of robotic systems. Predicting physical cause-and-effect relationships is essential for a robot’s performance in PRI tasks. Without this understanding, an agent cannot effectively distinguish between promising and unfavourable candidate actions, hindering its ability to perform tasks efficiently and safely.

    \begin{figure}[t!]
        \centering
        \includegraphics[width=0.7\textwidth, trim={0 62cm 0 0cm},clip]{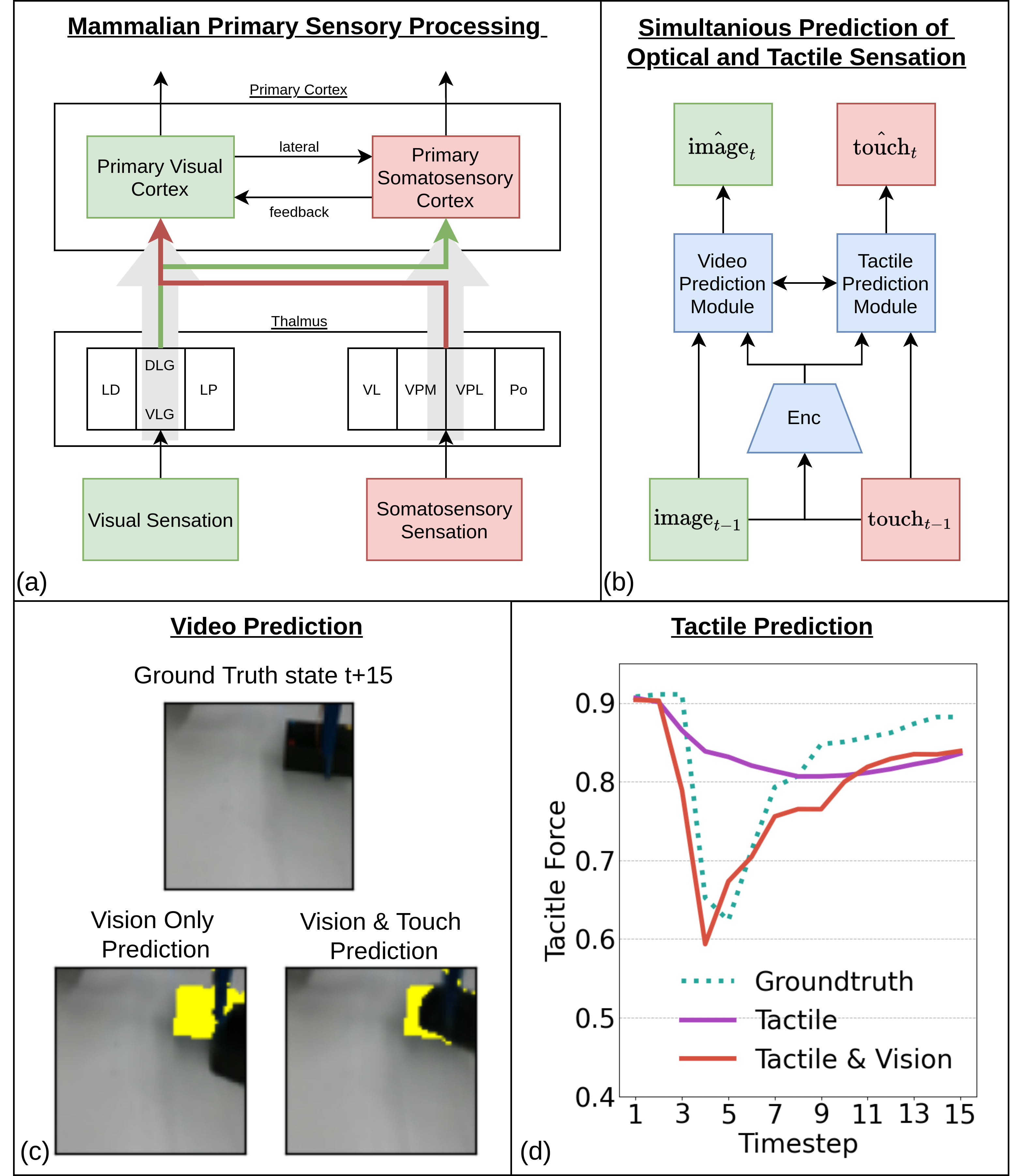}
        \caption{(a) Integration of touch and sight in the mammalian brain, involving cross-connections between the thalamus, primary somatosensory cortex, and visual cortex, which facilitates the fusion of tactile and visual information~\cite{Henschke2014Possible}; (b) Our best-performing model emulates this biological process, enhancing physical robot interactions by simultaneously predicting and integrating tactile and optical sensations.}
        \label{fig:abstractImage}
    \end{figure}
 
Visual and tactile sensations are fundamental to building physical interaction perception in humans~\cite{johansson2009coding}. Human tactile cognition, in particular, plays a crucial role in a variety of interactive tasks~\cite{nicholas2010active}, such as grasping, object manipulation, in-hand manipulation, tactile exploration, object pushing, and human-to-human physical collaboration. Research by Tseng et al.~\cite{tseng2007sensory} and Thoroughman et al.~\cite{thoroughman2000learning} demonstrates that humans rely on predictive models to execute these complex physical interaction tasks effectively.

In robotics, physical interaction tasks are typically handled by estimating physical property ~\cite{mavrakis2020estimating}, using interactive planning~\cite{mghames2020interactive}, learning based planning~\cite{tafuro2022dpmp} or deep neural network video prediction models~\cite{oprea2020review}. The latter ones are often benchmarked with object-pushing datasets, such as the BAIR dataset~\cite{ebert2017self}. More recently, tactile predictive models~\cite{nazari2022proactive} are studied and show superior performance compared to reactive tactile systems in physical robot interaction tasks. These prediction architectures utilize either optical or tactile sensations to perceive the environment and determine the control actions. However, unlike the multi-modal systems humans use for environment perception, these single-modality approaches result in a higher number of latent variables and increased prediction uncertainty. We propose that integrating both touch and vision into a multi-modal physical interaction perception model will improve an agent's cause-and-effect predictions during PRI tasks across both modalities. \chadded[id=AE]{To address this limitation, we propose a dual-pipeline multi-modal prediction architecture, termed SPOTS, which explicitly preserves modality-specific inductive biases while enabling cross-modal interaction.}

\begin{figure}[t!]
    \centering
    \includegraphics[width=0.98\textwidth, trim={0 0 0 0},clip]{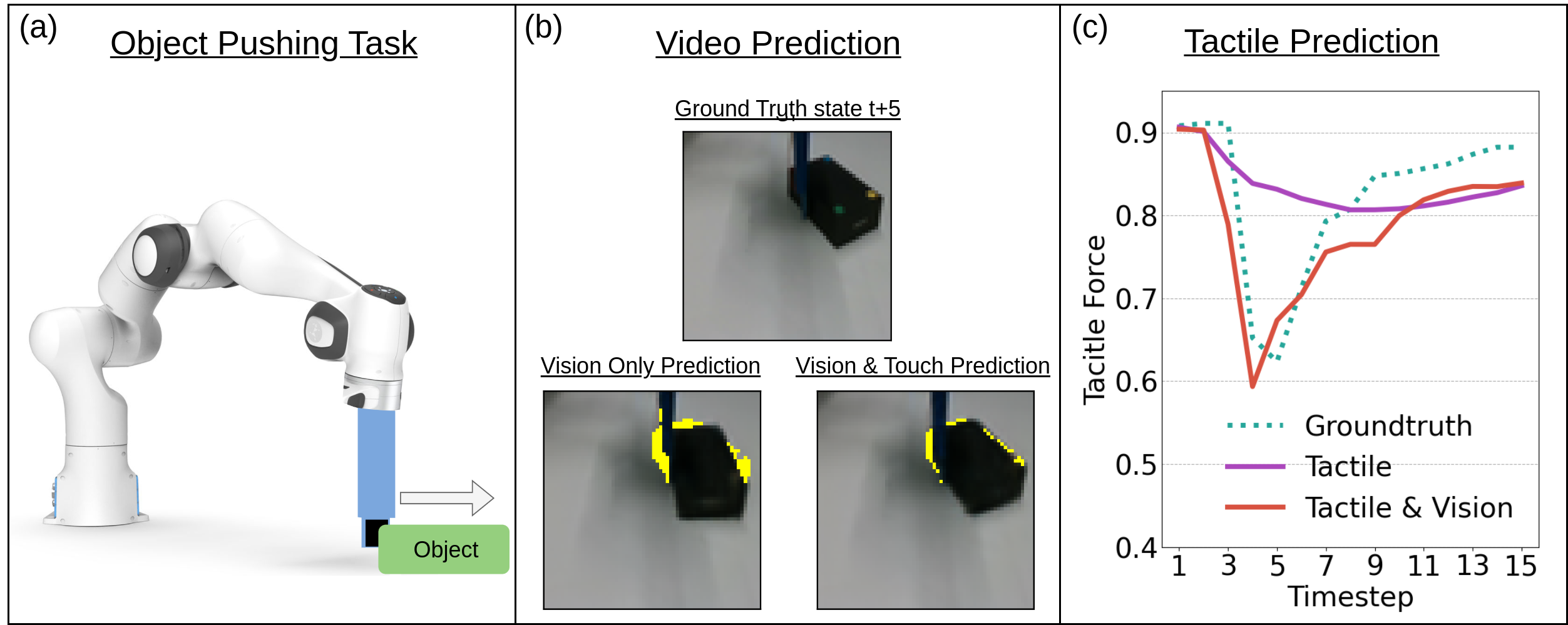}
    \caption{(a) The task we focus on is object pushing using the Franka Emika Panda; (b) Improved visual prediction accuracy, illustrated by the successful forecasting of a novel object's future position with complex physical properties during tactile-assisted robot pushing tasks. The yellow mask highlights the object’s actual future state at +5 time steps (0.5 seconds); (c) Our multi-modal architecture outperforms state-of-the-art tactile-only systems, as demonstrated by precise force predictions using a magnetic-based tactile sensor over extended prediction horizons.}
    \label{fig:abstractImage2}
\end{figure}

This work aims to explore and develop predictive physical perception models that leverage both visual and tactile sensations. In the context of human neuro-cognitive science~\cite{wolpert2001motor}, these models resemble the forward models employed in physically interactive tasks. In robotics, such models are implemented as deep neural networks that utilise the history of sensory readings, robot states, and planned future robot movements to generate predicted sensory readings over a prediction horizon. 
\chadded[id=AE]{Recent surveys and approaches~\cite{zhang2025step, guo2025ctrl, huang2026pointworld} on world models for robotic manipulation highlight the importance of learning predictive representations that capture physical dynamics across sensing modalities, particularly for contact-rich and partially observable interactions, further motivating the need for concrete, modality-aware visuo--tactile prediction architectures such as the one studied here.}

\chdeleted[id=AE]{Our best proposed model, \textbf{S}imultanious \textbf{P}rediction of \textbf{O}ptical and \textbf{T}actile \textbf{S}ensations (\textbf{SPOTS}), outperforms other state-of-the-art prediction architectures. It uses a dual pipeline prediction architecture that enables two bespoke network architectures dedicated to the prediction of the individual sensation. Crossover connections between the two pipelines capture the correlation between tactile and optical sensations and enable multi-modal learning. This bio-inspired approach mimics the structure and interaction between the visual and somatosensory primary cortexes in the mammalian brain (Fig. \ref{fig:abstractImage}), where we have individual cortexes for processing a given sensing modality, but crossover between the cortexes enables cross-sensation processing~\cite{Henschke2014Possible}.}

\chadded[id=AE]{To this end, we introduce SPOTS (Simultaneous Prediction of Optical and Tactile Sensations), a bio-inspired dual-pipeline \emph{world model} architecture (Fig. \ref{fig:abstractImage}) that explicitly models cross-modal interactions while preserving modality-specific inductive biases~\cite{Henschke2014Possible}. Rather than aiming for uniform improvements across all scenarios, SPOTS is designed to resolve latent physical properties—such as friction and contact dynamics, that cannot be inferred from vision alone.} 

\chadded[id=AE]{In this work, we investigate when and how tactile sensation meaningfully improves predictive perception in physical robot interaction. We show that while simple tactile conditioning provides limited benefit in visually unambiguous scenes, jointly predicting vision and touch becomes critical when physical properties are visually indistinguishable or when uncertainty accumulates over time.}

\chadded[id=AE]{Through extensive experiments on two tactile-visual robot-pushing datasets, including challenging edge cases with visually identical objects and altered physical properties, we demonstrate that SPOTS improves robustness, long-horizon prediction accuracy, and tactile prediction quality compared to both uni-modal baselines and simpler multi-modal fusion approaches. Our contribution includes: (1) We propose a dual-pipeline predictive architecture that enables simultaneous visual and tactile forecasting with cross-modal interaction. (2) We introduce two tactile-visual robot-pushing datasets, including a novel visually identical object dataset designed to isolate physical ambiguity. (3) We demonstrate that tactile-visual synergy provides limited benefit in visually unambiguous scenarios but yields significant gains in physically ambiguous and long-horizon prediction settings. (4) We show that SPOTS improves tactile prediction accuracy, generalisation to unseen physical properties, and robustness under sensory occlusion compared to state-of-the-art uni-modal and single-pipeline baselines.}

Recent research on combined visual and tactile sensations has focused on task-specific applications, where a multi-modal feature vector is used for tasks such as classifying grasp success~\cite{cui2020self}, identifying product defects~\cite{agarwal2023robotic}, translating between material properties~\cite{lee2019touching, cai2021visual}, or sensing modalities~\cite{li2019connecting}, object classification~\cite{wang2023human}, and object categorization~\cite{pinto2016curious}. For instance, Lee et al.~\cite{lee2020making} combined a force-torque sensor with visual features for a peg-in-the-hole reinforcement learning task. Overall, research is increasingly recognising the need for multi-modal approaches to solve complex robot interaction tasks. However, forward predictive models integrating these modalities have not yet been thoroughly explored.

We study the importance of integrating vision and tactile modalities by combining video prediction and tactile prediction architectures, specifically, the Action-Conditioned Tactile Prediction model~\cite{mandil2021tac} and the Stochastic Video Generator~\cite{denton2018stochastic} using various approaches~(Fig.~\ref{fig:abstractImage2}). We evaluate these methods in the context of object pushing, where the prediction system is tasked with forecasting future image and tactile frames of a scene based on previously observed frames and known robot actions. We believe that our unsupervised tactile-visual learning approach could benefit other aspects of physical-robot interaction, including grasping, in-hand manipulation, human-robot interaction, and soft tissue manipulation.

%%%%%%%%%%%%%%%%%%%%%%%%%%%%%%%%%%%%%%%%%%%%%%
\section{Related Works}    

In this section, we describe relevant research in video prediction and tactile prediction to provide an overview of existing uni-modal methods. We then discuss the literature surrounding physical robot interaction and the rationale for exploring our work in the context of robot pushing. Finally, we review existing multi-modal approaches. Although visuo-tactile prediction has not been extensively researched, the combination of video and tactile sensations has been explored in object recognition, robot control, physical interaction systems, and modality translation tasks, which serve as inspiration for our architectures.

Video prediction, which involves forecasting future video frames, is a fundamental challenge in enhancing robotic systems to perform human-like manipulation tasks. Early video prediction models focused on predicting raw pixel intensities without modelling scene dynamics~\cite{ranzato2014video}. To extend predictions over longer time horizons, Srivastava et al.~\cite{srivastava2015unsupervised} introduced autoencoders and LSTM units to capture temporal coherence. Action-conditioned video prediction models enhance predictions by incorporating additional information about actions~\cite{finn2016unsupervised, oh2015action}, facilitating reinforcement learning and model predictive control~\cite{ebert2017self}. Villegas et al.~\cite{villegas2017decomposing} approached the problem by splitting frames into content and motion streams, predicting the pose and dynamics of landmarks~\cite{villegas2017learning}. 

More recent methods have introduced stochastic assumptions in video prediction, recognising that multiple outputs can correspond to a single input due to latent variables. To address uncertainty, which often manifests as image blur~\cite{babaeizadeh2017stochastic}, models estimate and sample from latent variables to produce sharper predictions. Babaeizadeh et al.~\cite{babaeizadeh2017stochastic} applied this approach to the optical flow method proposed by Finn et al.~\cite{finn2016unsupervised}. Denton et al.~\cite{denton2018stochastic} proposed a simpler model with basic layers, which was later extended by Villegas et al.~\cite{villegas2019high} for high-fidelity video prediction. Lee et al.~\cite{lee2018stochastic} incorporated adversarial training techniques into the method from~\cite{finn2016unsupervised}.

Recent advancements have shifted to transformers, which use self-attention mechanisms to understand image structure by dividing images into patches. Video prediction, however, requires both spatial and temporal understanding. This has led to several approaches: Spatio-Temporal Transformers~\cite{bertasius2021space, arnab2021vivit}, Factorised Embedding Transformers~\cite{arnab2021vivit}, Factorised Self Attention Transformers~\cite{ye2022vptr, arnab2021vivit}, and Factorised Attention Transformers~\cite{arnab2021vivit}. Although these models offer competition with stochastic methods, they require large datasets for convergence. Collecting large-scale physical interaction datasets is very costly. Therefore, we selected an LSTM-based method as our baseline for comparison to demonstrate the performance of \emph{video and tactile prediction models} benefits from cross-modal integration (which is our main hypothesis). 

\subsection{Tactile Sensation technology}
Tactile sensors capture tactile information through physical interaction with the environment, measuring attributes such as temperature, vibration, softness, texture, shape, composition, and normal and shear forces~\cite{tiwana2012review,mandil2023tactile}. The technologies include acoustic-based tactile sensor~\cite{parsons2024single,mandil2024acoustic}, magnetic-base~\cite{xelauskin} and image based~\cite{yussof2010sensorization}, just to name afew. For object-pushing tasks, normal and shear force features are particularly important. Tactile sensors available in industry and literature typically balance resolution, affordability, and sensitivity.

\emph{Image-based} tactile sensors, such as optical waveguide-based sensors~\cite{yussof2010sensorization, ohka2005sensing} and marker-based sensors like the TacTip~\cite{winstone2012tactip}, provide high-resolution data but require significant processing~\cite{shah2021design}. 

\emph{Magnetic-based} sensors, such as the Xela uSkin~\cite{xelauskin}, provide low spatial resolution but high-frequency data at each Taxel (a single magnetic sensing element) with tri-axial readings. The Xela uSkin sensor measures non-calibrated normal and shear forces. Magnetic sensor sensors, e.g. Xela uSkin, are simple, low cost, and able to generate high-frequency readings. This sensor has been utilized in tactile predictive models~\cite{mandil2021tac} and for data-driven model predictive control in slip-free robotic manipulation~\cite{nazari2025bioinspired, nazari2022proactive}.

%\subsection{Tactile Prediction}
To predict tactile images from image-based tactile sensors, researchers have applied existing video prediction models. Tian et al.~\cite{tian2019manipulation} used the vision-based GelSight sensor~\cite{yuan2017gelsight} with an RNN-based prediction model, Convolutional Dynamic Neural Advection (CDNA)~\cite{finn2016unsupervised}, to propose a tactile-based model predictive control system for rolling an object to a target location under constant finger pressure.

For predicting magnetic-based tactile readings, Zhou et al.~\cite{zhou2020learning} converted Xela uSkin tactile sensor readings into a visual representation (an image with red dots representing taxel force values) to use with ConvLSTMs. However, this approach reduces the resolution of tactile readings and introduces issues with overlapping taxel dots, making interpretation challenging.

Zhang et al.~\cite{zhang2020towards} proposed an improved grasping system using a new video prediction model called PixelMotionNet, applied to tactile images from the FingerVision tactile sensor~\cite{yamaguchi2016combining}. This work aimed to improve grasp success rates by predicting contact and slip events, while our focus is on manipulation.

Most recently, Mandil et al.~\cite{nazari2021tactile, mandil2021tac} presented the Action-Conditioned Tactile Prediction (ACTP) network, which uses the Xela uSkin sensor to predict tactile signals over a short time horizon for slip prediction. This work found that vector-based LSTM layers significantly outperformed video prediction architectures like SVG and ConvLSTMs. The ACTP model has been used for adaptive real-time robotic control of slipping objects~\cite{nazari2022proactive} and clustered object pushing~\cite{nazari2023deep}.

Tactile prediction research is less developed compared to video prediction due to its narrower applicability outside physical robot interaction. However, evidence suggests that vector-based prediction methods outperform image-based methods for magnetic-based tactile sensation data. The success of predictive tactile models over-reactive systems in PRI tasks~\cite{nazari2022proactive} underscores the need for multi-modal prediction control architectures, suggesting future research directions for accurate multi-modal forward perception models in complex PRI tasks.
\subsection{Physical Robot Interaction}

Opera et al.~\cite{oprea2020review} categorize video prediction model benchmarks into three types: (i) human prediction tests~\cite{Ionescu6682899, schuldt2004recognizing, bregonzio2009recognising}, (ii) driving and road tests that predict changes in road states~\cite{brostow2009semantic, geiger2013vision, huang2018apolloscape}, and (iii) robot pushing datasets~\cite{finn2016unsupervised, ebert2017self, dasari2019robonet} where the objective is to predict environmental changes resulting from a robot's actions during physical interactions. Unlike other video prediction benchmarks, robot-pushing datasets involve physical interaction, and tactile sensation can provide features crucial for video prediction that are absent in visual data alone. State-of-the-art methods in video prediction often apply stochastic assumptions, treating variables such as the center of mass, object friction, and object dynamics as unknown. Tactile sensation during physical interaction can reveal many of these latent variables, motivating the integration of tactile sensation into video prediction models. 

%\subsection{Combining Vision and Touch}

The integration of vision and touch is still an emerging field. Research on combining vision and touch during physical robot interaction (PRI) has primarily focused on translation tasks. For instance, Lee et al.~\cite{lee2019touching} and Cai et al.~\cite{cai2021visual} used adversarial networks to translate between material surfaces and touch data, employing a vision-based touch sensor and a pen accelerometer, respectively. Early approaches often treated vision and tactile sensation as separate modalities within broader systems. For example, Agarwal et al.~\cite{agarwal2023robotic} used hierarchical classification to identify object defects, integrating vision and touch in a system for defect detection.

Typically, combining vision and touch aims to produce a multi-modal feature vector for specific tasks, such as control policies or task-specific features like object slip. Cui et al.~\cite{cui2020self} encoded visual and tactile features separately before using a self-attention network to generate a multi-modal feature vector, which was then applied to classify successful robot grasp positions. This approach outperformed a ResNet-encoded network combined with a Multi-Layer Perceptron (MLP)~\cite{calandra2018more}.

Li et al.~\cite{li2019connecting} employed ResNet encoders and adversarial training to (i) synthesize plausible temporal tactile signals from static visual inputs and (ii) translate tactile signals into a single image output of the scene. Lee et al.~\cite{lee2020making} combined vision, haptic data (wrist force/torque sensor), and proprioceptive data to create a multi-modal representation used for reinforcement learning in a peg-in-hole task, demonstrating that multi-modal representation outperforms single-modality models.

Li et al.~\cite{li2022see} fused acoustic, visual, and tactile data using ResNet to encode individual features, which were then concatenated and processed through a multi-headed self-attention layer. The system used an MLP to predict the next action for an occluded dense packing task.

Wang et al.~\cite{wang2023human} also fused vision and tactile sensation for object classification. Their Siamese system processes each modality separately using CNNs for images and LSTMs for tactile data. The modalities are fused via weighted concatenation and linear layers, with object classification based on the similarity between the two Siamese networks. Pinto et al.~\cite{pinto2016curious} demonstrated how physical interactions, such as pushing, poking, and grasping, can enhance object classification and categorization.

To date, the literature lacks models that capture the correlation between visual and tactile sensing to predict visual \emph{and/or} tactile sensing during PRI. We summarize potential methods for combining tactile and video prediction models to improve PRI task performance (see Fig.~\ref{fig:potentialapproaches}).

Overall, while vision and touch have been used in robot control, they are typically combined to generate a multi-modal feature vector for specific tasks, such as classification or action prediction. In contrast, our work explores forward physical interaction perception models that predict future states of the environment based on current and past readings. This approach aims to create a general multi-modal prediction framework with broader applicability, rather than focusing on task-specific solutions.
 
We identify a significant gap in integrating tactile and visual prediction domains to enhance prediction accuracy. We address the need for datasets that include tactile sensation, as existing pushing datasets lack this component. Moreover, our research will explore various prediction models, some inspired by the multi-modal fusion methods discussed, to answer our research question that `Does \emph{combined vision and tactile World Model} models outperform uni-modal one in physical robot interaction tasks?'. 

\section{Problem Formulation}

\begin{figure}[t!]
    \centering
    \includegraphics[height=0.3\textwidth]{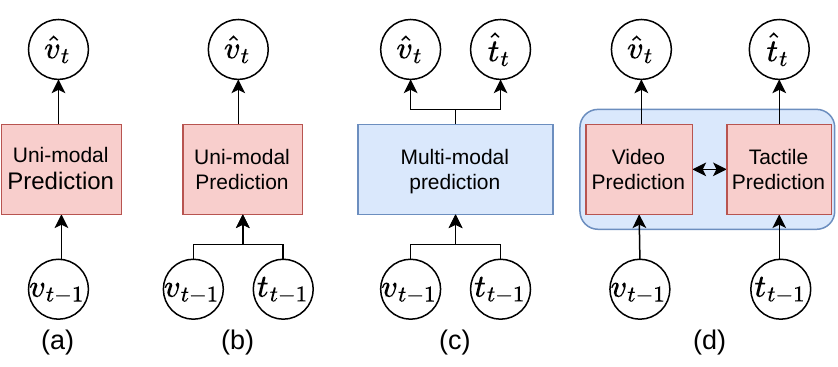}
    \caption{Possible methods of integrating tactile data into video prediction systems: (a) standard single-modality prediction; (b) incorporating a second modality as context to the uni-modal prediction model; (c) predicting both touch and vision sensations with a single prediction model; (d) using two separate prediction modules for each sensing modality, with a crossover link.}
    \label{fig:potentialapproaches}
\end{figure}
This work aims to develop a multi-modal forward perception model that: (i) improves video prediction by integrating tactile sensation, and (ii) enhances tactile prediction using video data. We explore various multi-modal prediction architectures to identify the most effective forward physical perception system. Our approach begins with uni-modal frame prediction models and progresses to multi-modal architectures, using the uni-modal models as benchmarks for comparison.
\begin{figure}
    \centering
    \includegraphics[width=1\linewidth, trim={1cm 1.45cm 2cm .2cm},clip]{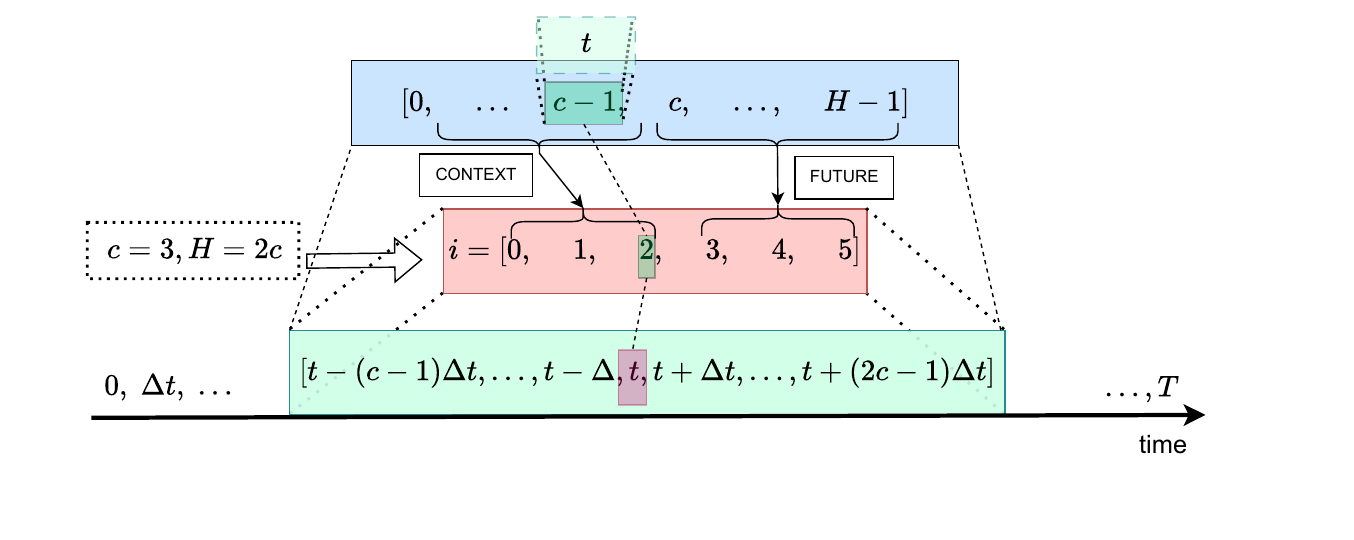}
    \caption{At each time $t$, a context and prediction window of size $c $ are used by the model where the entire context and prediction window has $H = 2 \times c$ time samples; hence, the context and prediction times are $\{t- c \Delta t, \dots, t- \Delta t, t\}$ and $\{t, t+ \Delta t, \dots, t+ (2c-1) \Delta t\}$ assuming fixed sampling frequency $\Delta t$ in dataset. For $c=3$, context is $\{t- 2 \Delta t, t- \Delta t, t\}$ and future frames $\{t, t+ \Delta t, t+ 2 \Delta t\}$ and the corresponding index in our prediction model is $i = \{0, 1, \dots, 5\}$~\cite{nazari2025bioinspired}.}
    \label{fig:time_prediction_time}
\end{figure}
\noindent \paragraph*{Video or Tactile Prediction} In this work, both video and tactile data are referred to as "frames", representing single time-step readings of the respective modality. Given: (i) a set of context frames $\textbf{f}_{0:c-1} = \{\textbf{f}_{0}, \dots, \textbf{f}_{c-1}\}$, which are the previously observed frames with a context sequence length of $c$, and (ii) a prediction horizon of length $H-c$ (that is the number of future frames to predict; here we assume equal context and prediction window size, i.e. $H = 2c$ where $i_c\in \{0, \dots, c-1\}$ and $i_p \in \{c, \dots, H-1\}$ where $i \in \{0, \dots, c-1\} \cup \{c, \dots, H-1\}$ (see Fig.~\ref{fig:time_prediction_time} for details of samples in the context and prediction horizon).  Although here we show the formulation for one time step $\textbf{f}(t)$, generalization for training for the entire trajectory, i.e. $t=0:T$, is straightforward.), a video or tactile prediction model can be defined as:\footnote{{\small We use $*_{i:i+n} = \{*_i, *_{i+1}, ..., *_{i+n} \}$, see Fig. \ref{fig:time_prediction_time} for detail}, $\textbf{f}$ to denote variables representing the set of frames (including both tactile and/or video), and $\hat{\textbf{f}}$ to denote the corresponding predicted values.}
    \begin{equation}
        \hat{\textbf{f}}_{c:H-1} = \mathcal{F}(\textbf{f}_{0:c-1}) 
        \label{eq:predictionmodel}
    \end{equation}
where $\hat{\textbf{f}}_{c:H-1}$ represents a set of predicted frames including video $\textbf{v}_i \in \mathbb{R}^{64 \times 64 \times 3}$ (\textit{scene images}) or tactile frames $\textbf{t}_i \in \mathbb{R}^{48}$ (\textit{tactile images}) or both, i.e. $\textbf{f}_i = \{\textbf{v}_i, \textbf{t}_i \}$. The goal is to optimize eq.~\ref{eq:predictioncost}, for each time step in the prediction horizon counted by $i_p$ (we consider equal length for context and prediction sequence length):
    \begin{equation}
        \min \sum_{i=c}^{H-1} \mathcal{D}\left( \hat{\textbf{f}}_i, \textbf{f}_i \right)
        \label{eq:predictioncost}
    \end{equation}
where $\mathcal{D}$ denotes the loss function in the tactile reading space or pixel space, such as $\mathcal{L}_1$ or $\mathcal{L}_2$, measuring the difference between predicted and observed frames.

\noindent \paragraph*{Action-Conditioned Video or Tactile Prediction} In physical robot interaction, we aim to develop a cause-effect understanding of the robot's actions. Thus, we condition the prediction model on the past context frames $\textbf{f}_{0:c-1}$, the past robot trajectory $\textbf{x}_{0:c-1}$, and planned robot actions $\textbf{a}_{c-1:H-2}$ to output future frames $\textbf{f}_{c:H-1}$ (which are known in our PRI datasets but unknown during inference time). Here, $\textbf{x}_{i_c} \in \mathbb{R}^7$ represents the robot trajectory at past steps, and $\textbf{a}_{i_p} \in \mathbb{R}^7$ represents the planned future robot actions at time $t$. The model assumes a known and nearly constant sampling frequency (typically around 10 Hz with less than 10\% variance), and we work with discrete values.
The prediction model can be expressed by wq.~\ref{eq:acpredictionmodel}.%
\begin{equation}
    \hat{\textbf{f}}_{c:H-1}= \mathcal{F}(\textbf{f}_{0:c-1}, \textbf{x}_{0:c-1}, \textbf{a}_{c-1:H-2})    \label{eq:acpredictionmodel}
\end{equation}
where $\textbf{f}$ can be video or tactile frames. 
In model predictive control (MPC), which commonly employs forward prediction models like those described here, future robot actions are considered as a batch of candidate actions. The optimal action is selected by a discriminator based on the most desirable predicted scene~\cite{nazari2022proactive, nazari2023deep}.
\paragraph{Simultaneous Video and Tactile Prediction} For predicting both modalities simultaneously, we denote video frames as $\textbf{v}_{0:H-1}$ and tactile frames as $\textbf{t}_{0:H-1}$. We describe the general approach for integrating these modalities here and will present specific model descriptions in the following sections.

Given the context frames for tactile and video, $\textbf{t}_{0:c-1} = \{\textbf{t}_{0}, \dots, \textbf{t}_{c-1}\}$ and $\textbf{v}_{0:c-1} = \{\textbf{v}_{0}, \dots, \textbf{v}_{c-1}\}$, respectively, and the past robot states and future actions $ \textbf{x}_{0:c-1} = \{\textbf{x}_{0}, \dots, \textbf{x}_{c-1}\}$ and $\textbf{a}_{c-1:H-2} = \{\textbf{a}_{c-1}, \dots, \textbf{a}_{H-2}\}$ respectively, the prediction model can be expressed as:
\[
p(\hat{\textbf{v}}_{c:H-1}, \hat{\textbf{t}}_{c:H-1} | \textbf{v}_{0:c-1}, \textbf{t}_{0:c-1}, \textbf{x}_{0:c-1}, \textbf{a}_{c-1:H-2}),
\]
where $\hat{\textbf{v}}_{c:T}$ and $\hat{\textbf{t}}_{c:T}$ represent the predicted video and tactile frames, respectively, over the prediction horizon $T-c$. This formulation allows the model to leverage both past and current multimodal sensory data and the robot's planned actions to simultaneously predict the future states of both video and tactile data, as per eq.~\eqref{eq:acpredictionmodel} where $\textbf{f}$ include both video and tactile frames.
The objective function for this model is:
\begin{equation}
    \min \sum_{i=c}^{H-1}  \mathcal{D}\left( \hat{\textbf{v}}_i, \textbf{v}_i \right)
    \label{eq:mmpredictioncost}
\end{equation}
\iffalse
We can condense equation \ref{eq:mmpredictionmodel} down by representing both the video $\textbf{v}_i$ and tactile $\textbf{t}_i$ frames as a set of frames $\textbf{f}_i = \{\textbf{v}_i, \textbf{t}_i \}$:
\begin{equation}
    \{ \hat{\textbf{f}}_{c:T} \}= \mathcal{F}(\textbf{f}_{0:c-1}, \textbf{x}_{0:c-1}, \textbf{a}_{c-1:t-1} ) 
    \label{eq:mmpredictionmodel2}
\end{equation}
\fi
This section provides a general problem formulation for multi-modal prediction. The following sections will detail various methods for integrating tactile and video data, starting with a simple tactile-conditioned video prediction model and advancing to single and dual pipeline multi-modal prediction models, as illustrated in Fig.~\ref{fig:potentialapproaches} (b, c \& d).

\subsection{Stochastic Video Generation} SVG model aims to maximize the $    p(\hat{\textbf{v}}_{c:H-1} | \textbf{v}_{0:c-1}, \{\textbf{x}_{0:c-1}, \textbf{a}_{c-1:H-2}\}, \textbf{z}_{0:c})$ to predict video images $\hat{\textbf{v}}_{c:T}$ by applying stochastic assumption to the prediction model~\cite{denton2018stochastic}. Likewise, our model also utilises stochastic assumption in the prediction model, where the objective is to sample from:
\begin{equation}
    p(\hat{\textbf{v}}_{c:H-1} | \textbf{v}_{0:c-1}, \{\textbf{x}_{0:c-1}, \textbf{a}_{c-1:H-2}\})
    \label{eq:sampleRNN}
\end{equation}
Within the base video prediction architecture, from which we build our models, video prediction is split into sub-modules: (i) a frame prediction network, (ii) a prior network, and (iii) a posterior network, which is used solely to train the prior network \cite{denton2018stochastic}. SVTG trades visual fidelity for improved physical localisation, explaining its lower pixel-level metrics but stronger qualitative performance.

In video prediction for Physical Robot Interaction (PRI), latent variables are utilized to estimate unknown physical properties: as an example, `when a robot’s arm pushes a toy on a table, the unknown weight of that toy affects how it moves~\cite{babaeizadeh2017stochastic}'. Intuitively, tactile sensation should provide the model with more accurate representations of the object's physical values. However, we continue using the stochastic assumption and latent variable estimation since other environmental features remain challenging to estimate even with tactile sensation.

We condition the frame prediction network on the estimated latent variables, $\textbf{z}$, so we now sample from:
\begin{equation}
    p(\hat{\textbf{v}}_{c:T} | \textbf{v}_{0:c-1}, \{\textbf{x}_{0:c-1}, \textbf{a}_{c-1:H-2}\}, \textbf{z}_{0:c})
    \label{eq:samplewlvRNN}
\end{equation}
The latent variables are distributed according to the prior network $q_{\psi}(\textbf{z}_i | \textbf{v}_{0:i-1})$ where $t$ is the current time step in the prediction sequence.
Learning involves training the parameters $\theta$ of the factorized model:
\begin{equation}
    \prod_{i=c}^{H-1} p_{\theta}(\hat{\textbf{v}}_{i} | \textbf{v}_{0:i-1}, \{\textbf{x}_{0:c-1}, \textbf{a}_{c-1:i-1}\}, \textbf{z}_{0:i})
    \label{eq:predictionmodel_deterministic}
\end{equation}
The learned prior network, $q_{\psi}(\textbf{z}_i | \textbf{v}_{0:i-1})$, is trained using Kullback-Leibler (KL) divergence \cite{kullback1951information} on the output of the posterior network $q_{\phi}(\textbf{z}_i | \textbf{v}_{0:i})$.
Both networks output the parameters of a conditional Gaussian distribution:
$$
\mathcal{N}(\mu_{\psi}(\textbf{v}_{0:i-1}), \: \sigma_{\psi}(\textbf{v}_{0:i-1}))$$
The prior network can then be trained jointly with the frame prediction model by maximizing eq.~\ref{eq:kdl}.
\begin{equation}
\begin{split}
    \mathcal{L}_{\theta, \phi, \psi}(\textbf{v}) & = \sum_{i=c}^{H-1}-  \big(\mathbb{E}_{q_{\phi}(\textbf{z}_i | \textbf{v}_{0:H-1})} \big[ \text{log} p_{\theta} (\hat{\textbf{v}}_{i:H-1} | \textbf{v}_{0:i-1}, \{\textbf{x}_{0:c-1}, \textbf{a}_{c-1:i-1}\}, \textbf{z}_{0:i}) \big] \\ 
    & + D_{KL} \big( q_{\phi}(\textbf{z}_i | \textbf{v}_{0:i}) || q_{\psi}(\textbf{z}_i | \textbf{v}_{0:i-1}) \big) \big) 
\end{split}
\label{eq:kdl}
\end{equation}
%\vspace{-.5cm}
For details on training the model and using the model at inference time see~\cite{denton2018stochastic}.
%%%%%%%%%%%%%%%%%%%%%%%%%%%%%%%%%%%%%%%%%%%%%%%
\begin{figure}[t!]
    \centering
    \subfloat[]{\adjustbox{fbox=0pt 4pt, frame}{\includegraphics[trim=0pt 0pt 0pt 0pt, clip, height=0.42\textwidth]{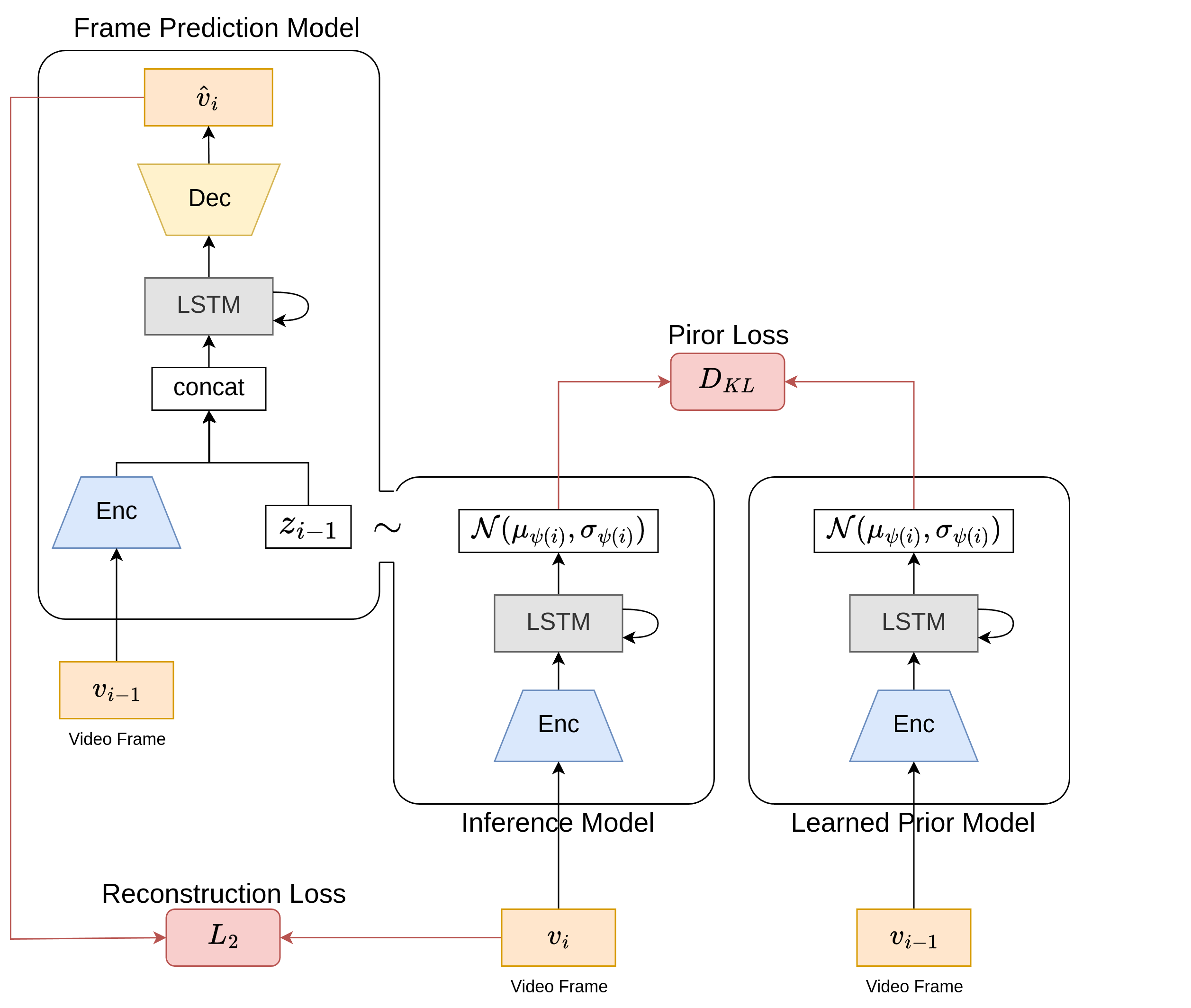}}}
    \hspace{0.1cm}
    \subfloat[]{\adjustbox{fbox=0pt 4pt, frame}{\includegraphics[trim=0pt 0pt 150pt 0pt, clip, height=0.42\textwidth]{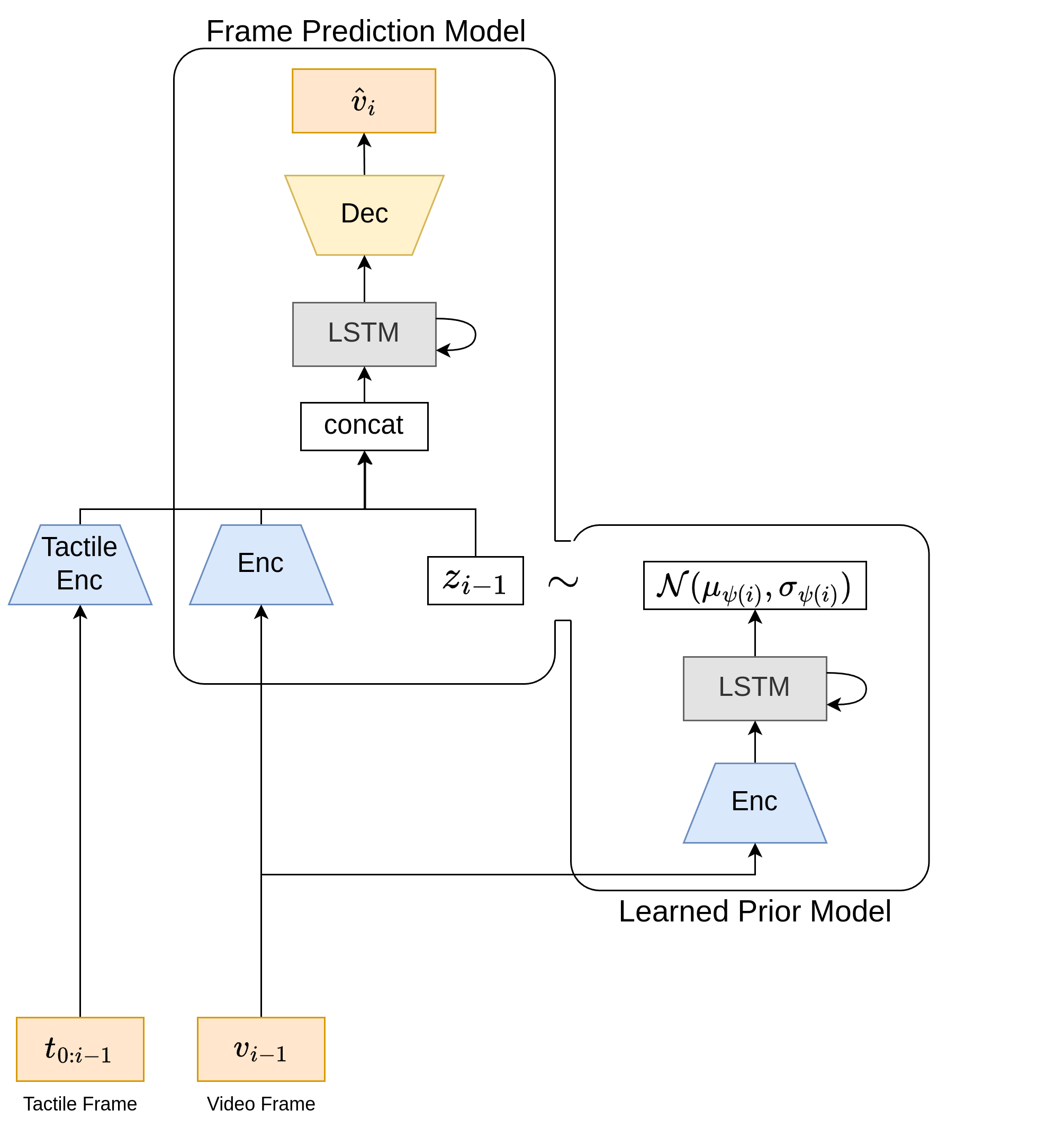}}}
    \caption{(a)  Stochastic Video Prediction Architecture SVG model without tactile enhancement for comparison~\cite{denton2018stochastic}, here we show the training structure which develops the Learned Prior Model which is utilized throughout our tactile enabled architectures; (b) SVG with integrated tactile sensation, called~\textbf{SVG-TE}, uses encoded tactile context data to enhance prediction accuracy. The diagram shows the test-time architecture.}
    \label{fig:models_svg_te}
\end{figure}
%%%%%%%%%%%%%%%%%%%%%%%%%%%%%%%%%%%%%%%%%%%%%%%
\begin{figure*}[t!]
    \centering
    \subfloat[]{\adjustbox{fbox= 0pt 4pt, frame}{\includegraphics[trim=0pt 0pt 150pt 0pt, clip, height=0.5\textwidth]{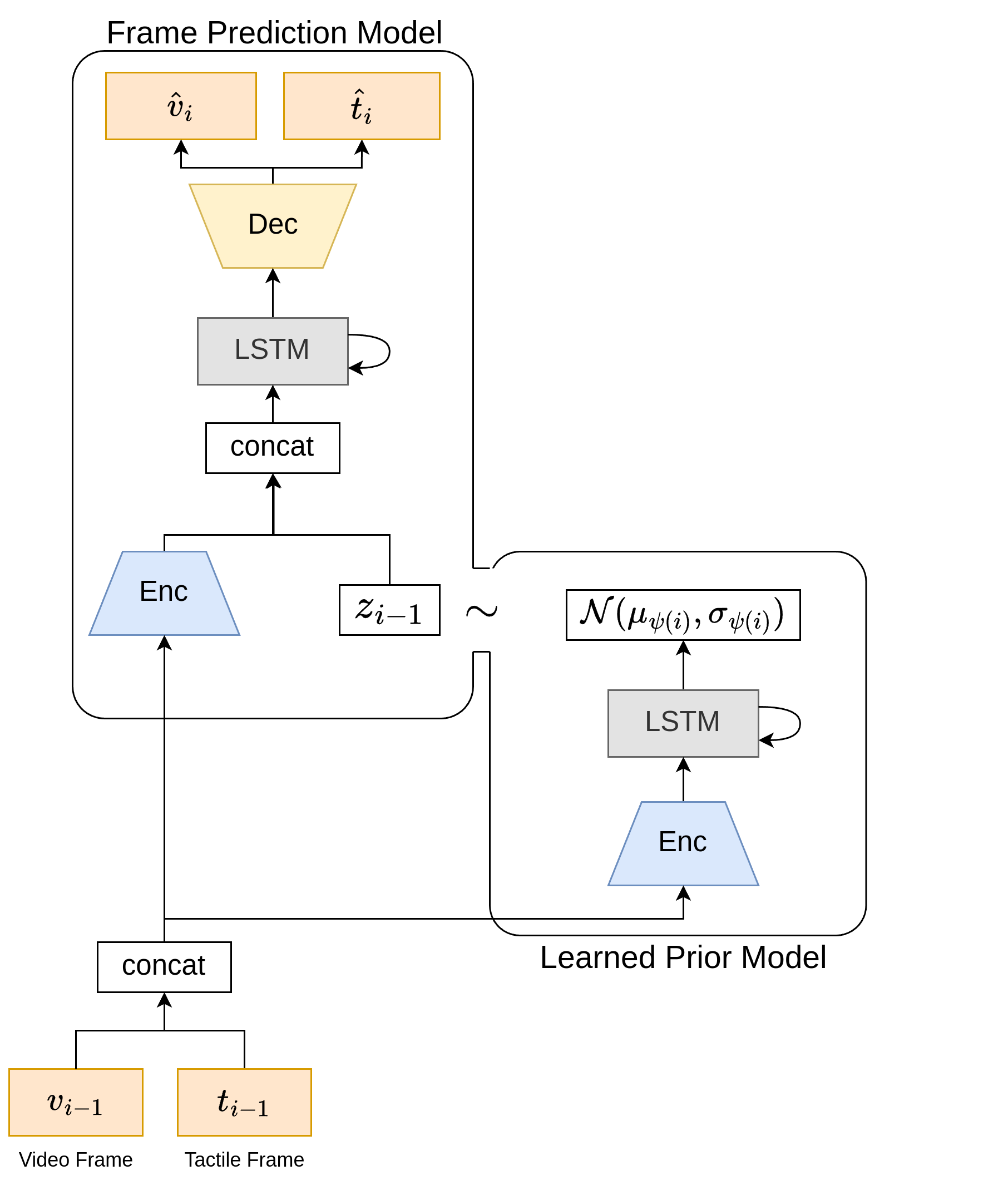}}}
        \hspace{0.1cm}
    \subfloat[]{\adjustbox{fbox= 0pt 4pt, frame}{\includegraphics[height=0.5\textwidth]{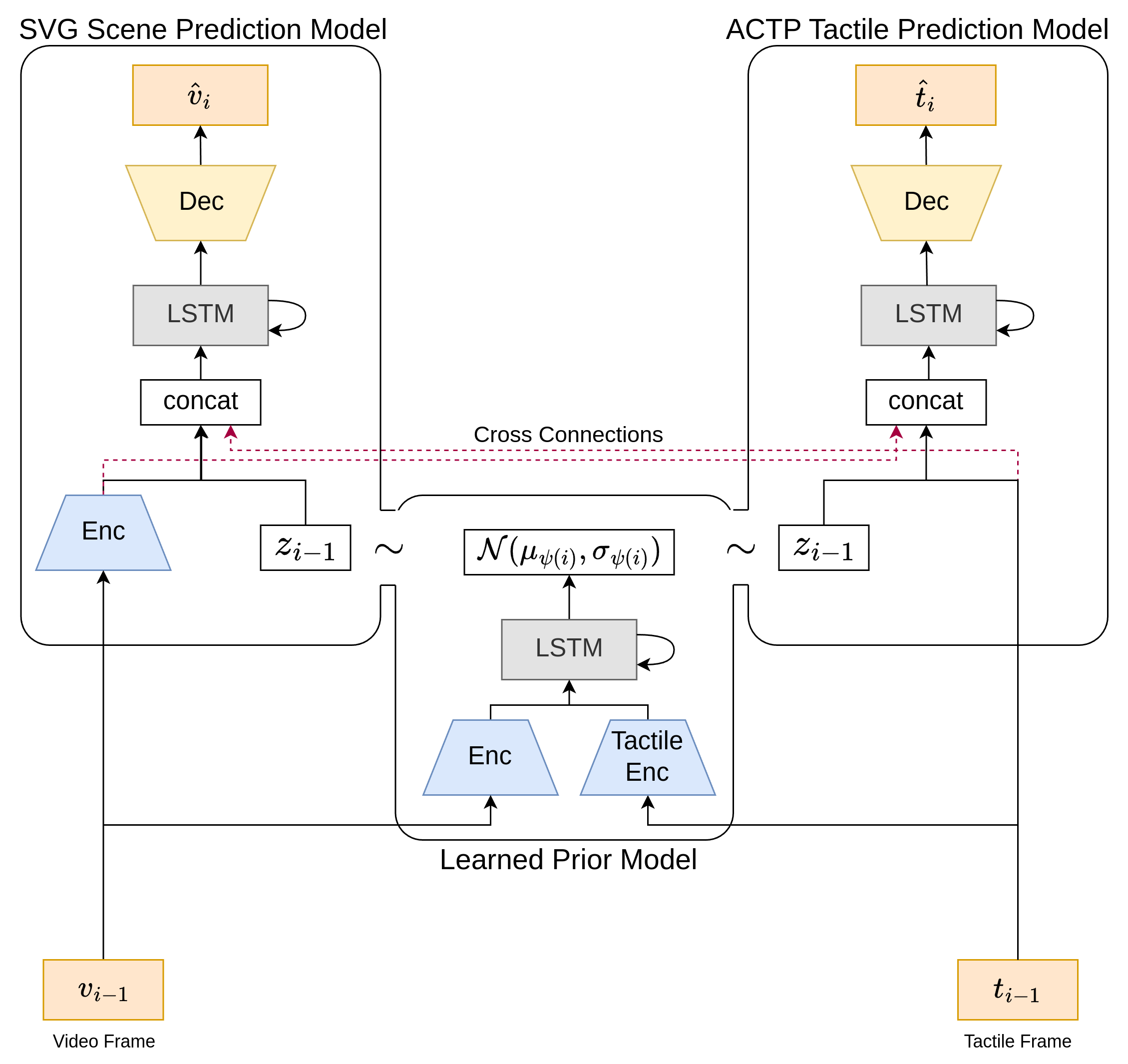}}}
    \caption{Multi-modal forward perception models based on the stochastic video prediction architecture SVG \cite{denton2018stochastic}. (a) \textbf{SVTG} (Stochastic Video and Tactile Generation) employs a single prediction pipeline to predict both vision and tactile sensations. (b) \textbf{SPOTS} (Simultaneous Prediction of Optic and Tactile Sensation) uses two separate prediction modules, one for each modality, with a crossover link, allowing for more accurate tactile prediction. We use the action-conditioned tactile prediction network ACTP \cite{mandil2021tac} for the tactile prediction pipeline.}
    \label{fig:models}
\end{figure*}
%%%%%%%%%%%%%%%%%%%%%%%%%%%%%%%%%%%%%%%%%%%%%%%%%%%%%%%%%%%%%%%%%%%%%%%%%%%%%%%

\section{Combining Vision and Tactile Sensation for Prediction}  
\label{section:method}
We chose the stochastic video prediction model SVG~\cite{denton2018stochastic} as the baseline architecture to build our multi-modal system from. SVG does not make the assumptions about the input data that other video prediction models like SAVP and SV2P do, making the system more generalisable \cite{villegas2019high} and hence more suitable to a multi-modal approach. Its simple architecture also allows us to modify the model's structure without a negative or destructive impact. 

While an auto-regressive transformer could also be a viable option due to its self-attention mechanism, which would enable effective cross-modality connections, we opted for an LSTM-based method. The primary reason is that our datasets are relatively small, and transformers typically require large amounts of data to perform optimally. The transformer's data-hungry nature might lead to sub-optimal results in our case, whereas LSTM-based methods are generally more data-efficient. To prove the hypothesis presented in this paper, either baseline network is acceptable, and a transformer-based version of this multi-modal prediction architecture is a potential direction for future work, particularly with larger datasets.
%\label{subsec:tactile_integration}
\subsection{Tactile Integration into Video Prediction Models} Various methods can be employed to integrate tactile sensations into video prediction models. In this section, we discuss the approaches to incorporating tactile information into the video prediction architecture, as illustrated in Fig.~\ref{fig:models_svg_te}.

It is important to distinguish between the different types of sensory data.
There are three primary approaches to integrating tactile sensation into video prediction models (see Fig.~\ref{fig:potentialapproaches}): (\emph{i}) conditioning the SVG model directly on context tactile data (Fig.~\ref{fig:models_svg_te}); (\emph{ii}) concatenating the scene and tactile data, and then feeding the combined data into a single prediction model to generate both touch and scene predictions (Fig.~\ref{fig:models}(a)); (\emph{iii}) utilizing separate prediction models for scene and tactile data, with cross-connections between the two modalities (Fig.~\ref{fig:models}(b)). 

The following sections provide detailed descriptions of these integration strategies, highlighting their design, key features, and the layers involved. We also outline our comparison study in Section~\ref{results}.
In all models, the robot's future action data, $\textbf{a}_{c-1:t-1}$, and past states, $\textbf{x}_{0:c-1}$, are concatenated~\cite{finn2016unsupervised} and then input into the LSTM chain of the Frame Prediction Model, along with other feature vectors.

\paragraph*{Tactile-Conditioned Video Prediction:} We compare the performance of multi-modal prediction against a video prediction model conditioned on past tactile frames. This comparison evaluates whether the multi-modal architecture can leverage predicted future tactile readings to enhance video prediction accuracy.

The simplest integration method involves flattening the past tactile images, $\textbf{t}_{0:c-1} = \{\textbf{t}_0, \dots, \textbf{t}_{c-1}\}$, where $\textbf{t}_t \in \mathbb{R}^{4 \times 4 \times 3}$, is converted into $\textbf{t}_t \in \mathbb{R}^{48}$. These flattened frames are then encoded into a feature vector and concatenated with the robot action data, the learned latent variables, and the scene feature vector. This composite input is fed into the LSTM chain of the Frame Prediction Model. The model, referred to as Tactile-Enhanced Stochastic Video Generation (SVG-TE), is depicted in Fig.~\ref{fig:models_svg_te}. Notably, in this architecture, the tactile feature vector does not contribute to the latent variable computation.

Training the SVG-TE model involves learning the weights of the frame prediction network ($\theta$), the learned prior network ($\psi$), and the posterior network ($\phi$). The final factorized frame prediction model is given by eq.~\ref{eq:SVGTEkdl}:
\begin{equation}
    \prod_{i=c}^{H-1} p_{\theta}(\hat{\textbf{v}}_{i} | \textbf{v}_{0:c-1}, \{\textbf{x}_{0:c-1}, \textbf{a}_{c-1:i-1} \}, \textbf{t}_{0:c-1}, \textbf{z}_{0:i})
    \label{eq:SVGTEkdl}
\end{equation}
And the loss function for learning $\theta$, $\psi$, and $\phi$ is given by eq.~\ref{eq:predictionmodeldeterministicSVGTE}.
\begin{equation}
\begin{aligned}
    \mathcal{L}_{\theta, \phi, \psi}(\textbf{v}) = &
    \sum_{i=c}^{H-1} \Bigg[ - \mathbb{E}_{q_{\phi}(\textbf{z}_{0:i} | \textbf{f}_{0:i})} \left( \log p_{\theta} (\textbf{v}_{i} | \textbf{v}_{0:i-1}, \{\textbf{x}_{0:c-1}, \textbf{a}_{c-1:i-1} \}, \textbf{t}_{0:i-1}, \textbf{z}_{0:i}) \right)  \\
    &+ D_{KL} \left( q_{\phi}(\textbf{z}_i | \textbf{v}_{0:i}, \textbf{t}_{0:i} ) \, \| \, q_{\psi}(\textbf{z}_i | \textbf{v}_{0:i-1}, \textbf{t}_{0:i-1}) \right) \Bigg]
\end{aligned}
\label{eq:predictionmodeldeterministicSVGTE}
\end{equation}

\subsection{Simultaneous Tactile and Video Prediction} The following two architectures are designed to predict both tactile and scene frames simultaneously. We adapt our model to sample from probability in eq.~\ref{eq:SVTG_full}
\begin{equation}
    p_{\theta}(\textbf{v}_{c:H-1}, \textbf{t}_{c:H-1} \mid \textbf{x}_{0:c-1}, \textbf{a}_{c-1:H-2} , \textbf{t}_{0:c-1}, \textbf{v}_{0:c-1},\textbf{z}_{0:c-1}),
    \label{eq:SVTG_full}
\end{equation}
as illustrated in Fig.~\ref{fig:potentialapproaches} (c) and (d). Our objective is now to minimise the following cost in eq.~\ref{eq:SPOTScost}:
\begin{equation}
    \min \sum_{i=c}^{H-1} \alpha \mathcal{D}\left( \hat{\textbf{v}}_i, \textbf{v}_i \right) + \beta \mathcal{D}\left( \hat{\textbf{t}}_i, \textbf{t}_i \right)
    \label{eq:SPOTScost}
\end{equation}
where $\alpha$ and $\beta$ are weightings for the modality loss functions; in our work, these are set to 1.0, though future research may explore different weightings. \chadded[id=AE]{Fixing $\alpha = \beta$ avoids the introduction of task-specific tuning and ensures that performance differences arise from architectural design rather than loss engineering.}

We factorise these models according to eq.~\ref{eq:SPOTSkdl}, where our objective translates into the loss $\mathcal{L}_{\theta, \phi, \psi}(\textbf{f})$ in SPOTS and SVTG with condensed notations by full model training.
\begin{equation}
    \prod_{i=c}^{H-1} p_{\theta}(\textbf{v}_{0:i}, \textbf{t}_{0:i}) | \textbf{v}_{0:c-1}, \{\textbf{x}_{0:c-1}, \textbf{a}_{c-1:i-1} \}, \textbf{t}_{0:c-1}, \textbf{z}_{0:i})
    \label{eq:SPOTSkdl}
\end{equation}

\iffalse
\begin{equation}
\begin{aligned}
    \mathcal{L}_{\theta, \phi, \psi}(\textbf{f}) = &
    \sum_{i=c}^{H-1} \Bigg[ - \mathbb{E}_{q_{\phi}(\textbf{z}_{0:i} | \textbf{f}_{0:i})} \left( \log p_{\theta} (\textbf{f}_{i} | \textbf{f}_{0:i-1}, \{\textbf{x}_{0:c-1}, \textbf{a}_{c-1:i-1} \}, \textbf{z}_{0:i}) \right)  \\
    &+ D_{KL} \left( q_{\phi}(\textbf{z}_i | \textbf{f}_{0:i} ) \, \| \, q_{\psi}(\textbf{z}_i | \textbf{f}_{0:i-1}) \right) \Bigg]
\end{aligned}
\label{eq:predictionmodeldeterministicSPOTS}
\end{equation}
\fi
We hypothesise that incorporating predicted tactile frames into the scene frame predictor network will enhance prediction performance beyond the improvement provided by context tactile data alone. Additionally, predicting tactile sensations enables more advanced applications in model predictive control scenarios, such as proactive slip control \cite{nazari2022proactive}.

\subsection{Proposed Combined Video and Tactile Prediction Models} 
We evaluate these hypotheses using the two proposed architectures (shown in Fig.~\ref{fig:models}~(a)), Stochastic Video and Tactile Generator (\textbf{SVTG}) and Fig.~\ref{fig:models}~(b) Simultaneous Prediction of Optic and Touch Sensations (\textbf{SPOTS}). The following sections discuss additional layers and key features of these models developed with this approach.
\emph{SVTG:} 
This architecture concatenates scene and tactile data before encoding, as shown in Fig.~\ref{fig:models} (a). The tactile data is reshaped from $\textbf{t} \in \mathbb{R}^{48}$ to $\textbf{t} \in \mathbb{R}^{64 \times 64 \times 3}$, where the three channels represent normal, shear x, and shear y forces\footnote{The Xela sensor used in this paper has 4 x 4 sensing cells, with each cell providing normal, shear x, and shear y readings at each taxel.}. Despite its simplicity, the SVG architecture has been found to predict rescaled tactile data poorly \cite{mandil2021tac}. To address this, we also implemented the SPOTS architecture, which is described below.

\emph{SPOTS:}
This model utilizes two Frame Predictor Models—one for each modality—as illustrated in Fig.~\ref{fig:models}~(b). The SVG's frame predictor model handles scene prediction, while a bespoke tactile prediction model, specifically the Action-Conditioned Tactile Prediction (ACTP) \cite{mandil2021tac}, is employed for tactile frame prediction. Crossover connections between the encoded tactile and scene data allow each pipeline to leverage information from the other modality.

We chose to split the prediction network into separate image and tactile pipelines for several reasons: (i) It allows for modifications to the model structure that can be specific to each modality, such as applying optical flow for video prediction but not for tactile prediction; (ii) This architecture facilitates the integration of additional modalities, which may require unique architectures, such as auditory or olfactory sensors \cite{Henschke2014Possible}; (iii) The split design makes it easier to adapt to specific domain issues—modifications to the scene prediction network do not affect tactile prediction performance, and vice versa.
Several adjustments were made to support this dual-pipeline approach.

\emph{Multi-Modal Fusion Model (MMFM):} Inspired by \cite{lee2020making}, we integrate the two sensing modalities using an MMFM layer. The MMFM consists of two linear layers with batch normalisation and \emph{tanh} activation functions. Since the optimal multi-modal representation may differ for each network modality, each pipeline in the SPOTS architecture has its own MMFM layer. This MMFM layer is positioned before the LSTM chain in each pipeline and processes the encoded scene and tactile values.

\emph{Scene-only Learned Prior:} The learned prior network generates latent variables that help reduce scene prediction blur by estimating latent values. This approach improves performance metrics that correlate with human-based image similarity scores, such as the Structural Similarity Index (\textbf{SSIM}) and Peak Signal-to-Noise Ratio (\textbf{PSNR}). However, it is unclear whether this is necessary for the tactile prediction network, where deterministic predictions might be more beneficial. We evaluate this option with \textbf{SPOTS-SOP} (SPOTS Scene-Only Prior).

\emph{Action-Conditioned Tactile Prediction Network:} The dual pipeline system allows us to tailor each pipeline to its modality, facilitating a more detailed exploration of integration challenges. While video prediction architectures like SVG can handle tactile data by resizing it, this approach has shown limited effectiveness. To address this, we employ the state-of-the-art tactile prediction model, ACTP~\cite{mandil2021tac}, in the tactile prediction pipeline.

%%%%%%%%%%%%%%%%%%%%%%%%%%%%%%%%%%%%%%%%%%%%%%%%%%%%%%%%%%%%%%%%%%%%%%%%%%%%%%%%
\begin{figure*}[t]
    \centering
    % Added a frame for better visibility and adjusted padding.
    \adjustbox{fbox=0pt 4pt, frame}{
        \includegraphics[trim=200pt 300pt 0pt 200pt, clip, height=0.4\textwidth]{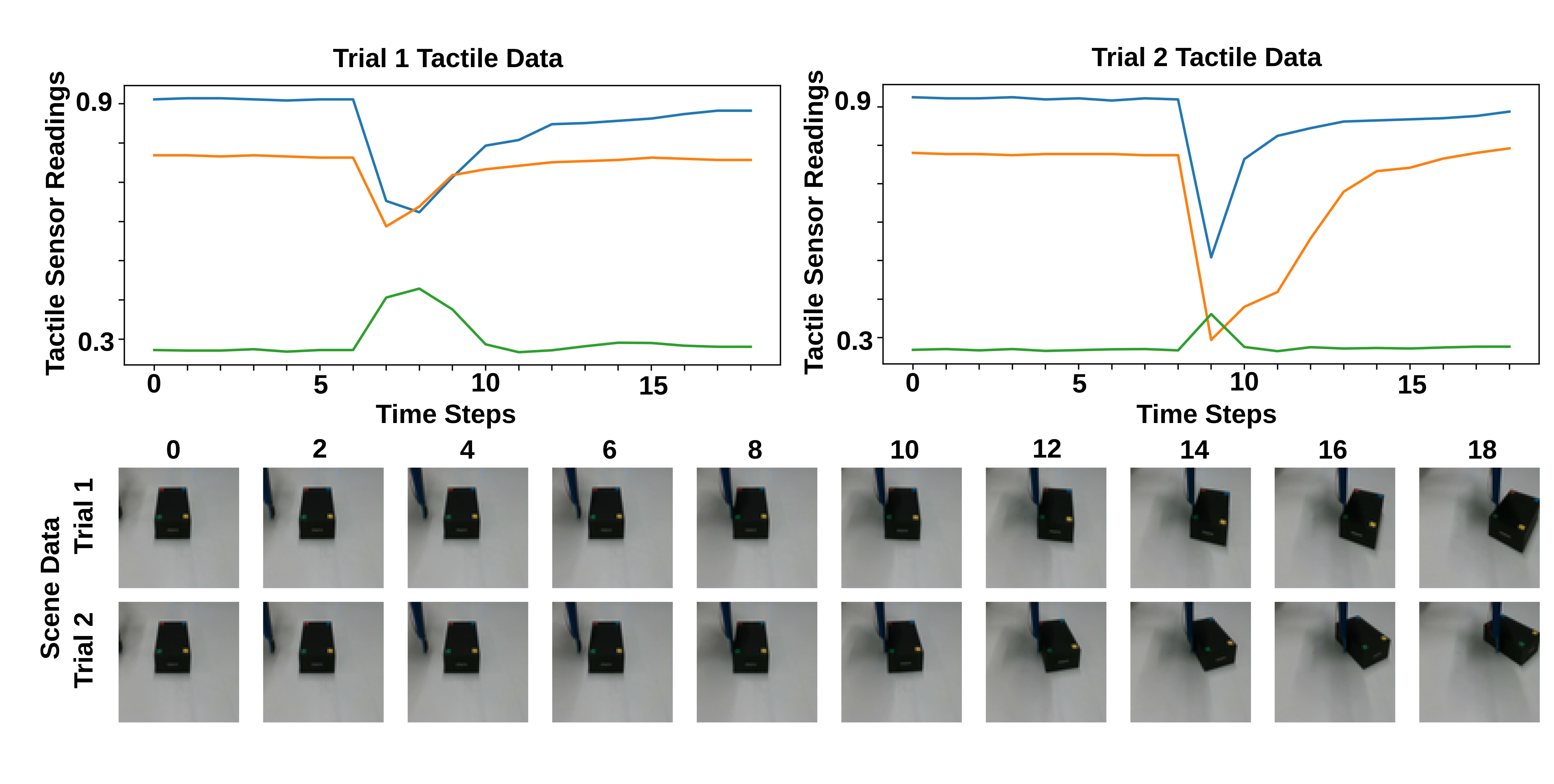}
    }
    \caption{
        Illustration of two trials from the edge case subset. Each trial presents scene video frames alongside three normalised taxel values: normal force, shear X, and shear Y. Despite starting from the same initial position, the final positions of the object differ due to varying friction locations in each trial. This setup presents a challenging scenario for the prediction agent, testing its ability to perceive and respond to complex physical interactions.
    }
    \label{fig:EdgedDataSeq}
\end{figure*}
%%%%%%%%%%%%%%%%%%%%%%%%%%%%%%%%%%%%%%%%%%%%
\section{Experiment Setup}
To train and evaluate the tactile-integrated models described above, we developed two novel object-pushing datasets incorporating both tactile sensations and scene videos. Object-pushing datasets are widely employed for benchmarking video prediction tasks \cite{oprea2020review} because they specifically assess physical robot interactions, making them ideal for evaluating our prediction models. This is in contrast to other benchmarks that focus on driving or urban scene understanding.

Previous research in Physical Robot Interaction (PRI) video prediction has primarily focused on datasets that assess generalisation across household objects and clusters \cite{dasari2019robonet, ebert2017self,finn2016unsupervised}. In line with these methodologies, we created our datasets by performing random robot-pushing actions on household object clusters. The datasets comprise: (i) robot proprioception data in both joint and task space, enabling action conditioning ($\textbf{a}_i \in \mathbb{R}^{7}$ and $\textbf{x}_i \in \mathbb{R}^{7}$)(ii) tactile data from the pushing finger of the gripper ($\textbf{t}_i \in \mathbb{R}^{48}$); and (iii) RGB-D video frames captured from three different perspectives of the scene ($\textbf{v}_i \in \mathbb{R}^{64 \times 64 \times 4}$). The synchronised data was recorded at a rate of 10 frames per second.

We introduced tactile sensation by attaching the Xela uSkin magnetic tactile sensor to the robot's pushing fingertip. The Xela uSkin tactile sensor features 16 sensing elements arranged in a \emph{4 by 4} grid, each providing non-calibrated readings of shear x, shear y, and normal forces. These readings are proportional to the corresponding forces. The Xela sensor has been previously employed for predicting tactile sensations in pick-and-move tasks \cite{mandil2021tac}, making it well-suited for our scene and tactile prediction models. Alternative vision-based sensors, e.g. the GelSight sensor \cite{yuan2017gelsight}, could also be employed in future works.

For the pushing tasks, we utilised the Franka Emika Panda robot, while scene frames were captured using the Intel RealSense D345 camera. Unlike previous robot-pushing datasets that involve random object pushes, our datasets consist of straight-line pushes, which offer two primary advantages: (i) they ensure consistent contact between the tactile sensor and the objects; and (ii) they facilitate continuous interactions with the objects over extended time horizons. For instance, during random motions, objects are often touched but not fully pushed through. In contrast, straight-line pushes maximise the displacement of objects, creating the complex scene dynamics necessary for rigorous model testing.

\subsection{Household Object Clusters Pushing Dataset}
\begin{figure*}[t]
        \centering
        \subfloat[]{\adjustbox{fbox= 0pt 4pt, frame}{\includegraphics[trim=0pt 0pt 0pt 0pt, clip, height=0.273\textwidth]{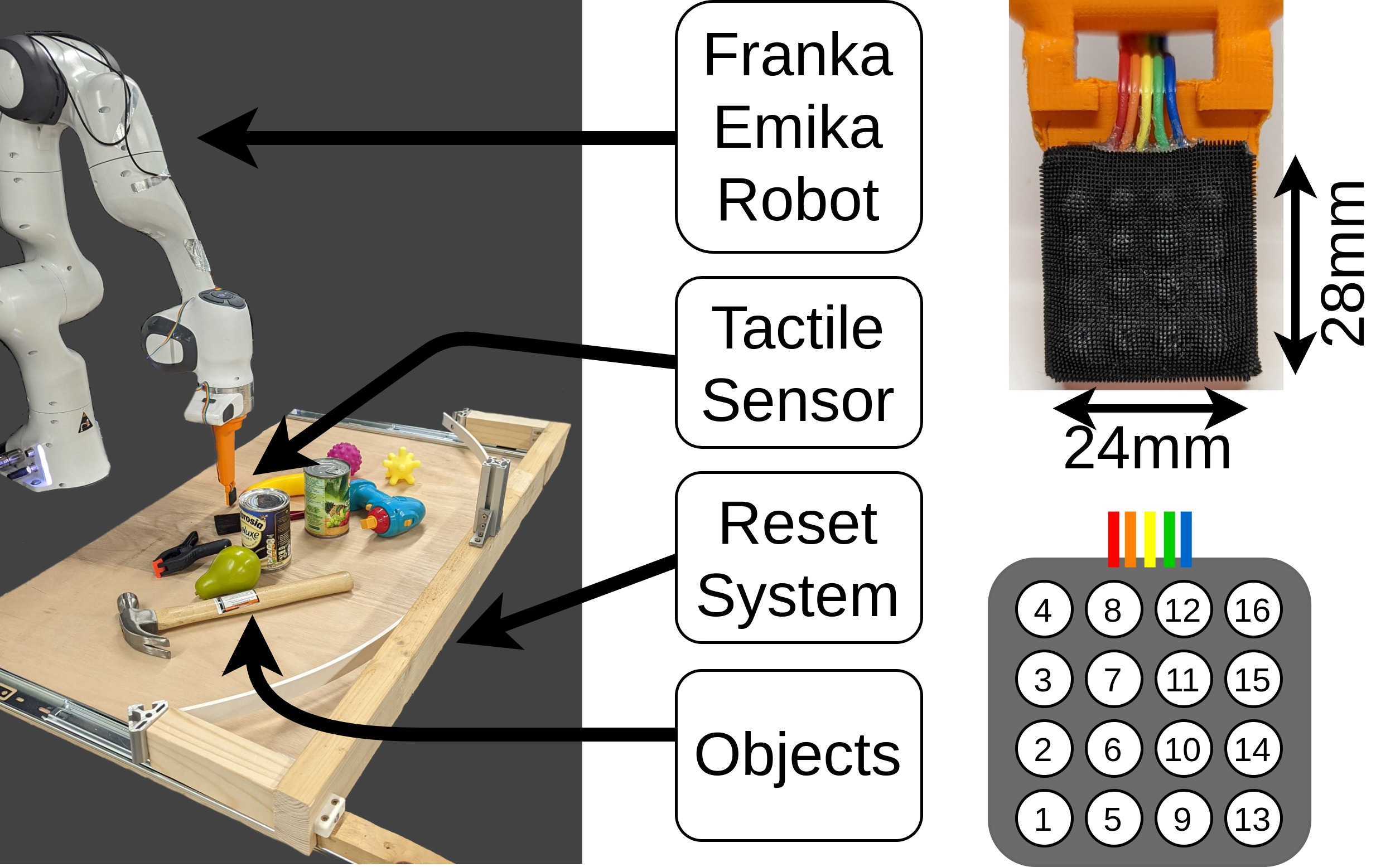}}}
        \hspace{0.1cm}
        \subfloat[]{\adjustbox{fbox= 0pt 4pt, frame}{\includegraphics[trim=0pt 0pt 0pt 0pt, clip, height=0.273\textwidth]{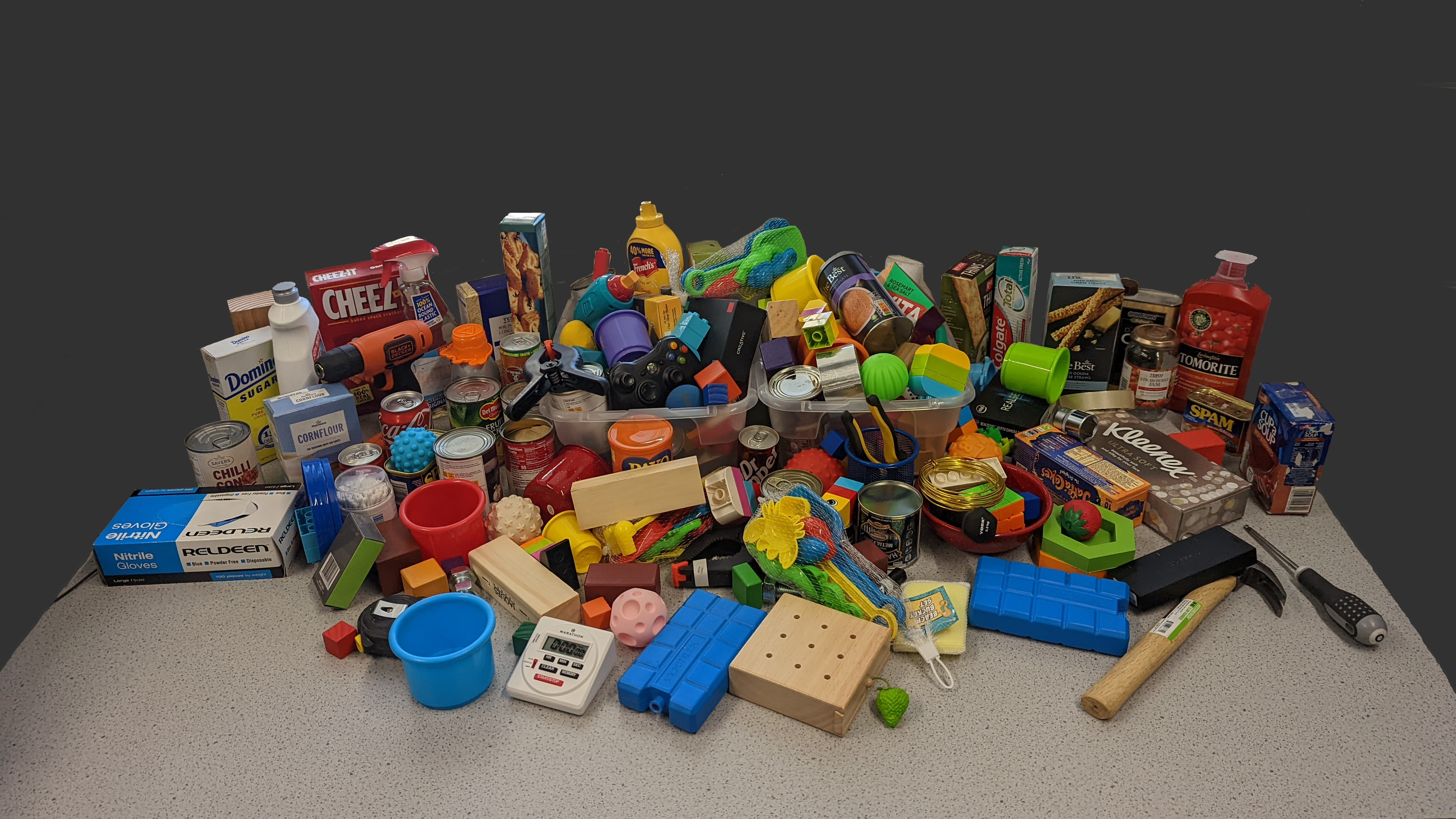}}}
        \caption{(a) The robot and its environment are shown, containing the Panda Franka Emika 7 degrees of freedom collaborative robot, the 4x4 Xela uSkin tactile sensor attached to the pushing fingertip, the household objects on the pushing surface and the object reset system, which enabled semi-automated dataset collection. (b) The objects used in the household object clusters dataset.}
        \label{fig:Datasetsetup}
    \end{figure*}

This dataset consists of 5,500 pushing trials involving clusters of hundreds of household objects (Fig.~\ref{fig:Datasetsetup}~(b)). The objects are primarily sourced from the YCB dataset \cite{calli2015ycb}, with additional items included to ensure robust generalisation testing. Each trial lasts for 4 seconds. The test data is divided into two sets: (i) \emph{seen object clusters}, comprising new clusters assembled from objects included in the training dataset, and (ii) \emph{unseen object clusters}, containing objects not present in the training dataset. Each test set includes 250 trials.
%%%%%%%%%%%%%%%%%%%%%%%%%%%%%%%%%%%%%%%%%%%%%%%%%
%%%%%%%%%%%%%%%%%%%%%%%%%%%%%%%%%%%%%%%%%%%%%%%%%

The dataset collection process was semi-automated. After every 20 pushes, an automated reset procedure was triggered, where the robot utilised an elliptical barrier to push all objects back to a central location (see Fig.~\ref{fig:Datasetsetup}). This method increased the likelihood of object contact in each trial. The setup is depicted in Fig.~\ref{fig:Datasetsetup}~(a). The arena featured boundaries beyond the pushing range, minimising instances where objects were pushed against barriers (typically, only the largest objects reached the edges).

\noindent \textbf{Visually Identical Dataset:} This dataset is designed to test the model's ability to integrate tactile data for scene understanding. It involves the same object across trials, but with friction markers placed at different locations on the contact surface. This setup specifically targets our task, enabling a qualitative assessment of the impact of tactile integration. Without tactile sensing, the model would likely fail to predict the correct motion direction of the object, as no visual cues indicate the high-friction areas.

The dataset includes 1,000 training interactions and 600 test interactions. The object used in this dataset is a single, heavy item (1.1 kg, dimensions: 16.1 × 10.1 × 4.9 cm), with friction modified by applying \emph{60-grit} sandpaper to various parts of the contact surface. These alterations lead to different interaction outcomes despite the same visual appearance (Fig.~\ref{fig:EdgedDataSeq}). The friction locations include the centre, the four corners, and the middle of the four edges of the box. The test set consists of 550 pushes with unseen friction locations and 50 pushes with previously seen friction locations. Colored markers were used to quantify object location and orientation during performance evaluation.

\begin{table}[t]
    \centering
    \begin{tabular}{llllll}
        \hline
        Model Name   & TE  & TP & SOP & MMFM & Model Size \\
        \hline
        SVG          & $\times$ & $\times$ & $\times$ & $\times$ & 18,010,027 \\
        SVG-TE       & \checkmark & $\times$ & $\times$ & $\times$ & 18,115,327 \\
        SVTG         & $\times$ & \checkmark & \checkmark & $\times$ & 38,766,766 \\
        SPOTS        & $\times$ & \checkmark & $\times$ & \checkmark & 21,051,051 \\
        SPOTS-Small  & $\times$ & \checkmark & $\times$ & \checkmark & 18,250,180 \\
        SPOTS-SOP    & $\times$ & \checkmark & \checkmark & \checkmark & 21,026,475 \\
        \hline
    \end{tabular}
    \caption{Key features of models tested in the comparison study, where: TE indicates a model performing tactile-enhanced scene predictions; TP indicates a model performing both tactile and scene prediction; SOP indicates a model with only scene data as input to the learned prior; MMFM indicates a model using a Multi-Modal Fusion Model layer. Model size refers to the number of parameters in the network. We test SPOTS-Small, which has the same number of parameters as the vision-only method, to ensure model size is not the cause of improved performance.}
    \label{tab:keyfeatures}
\end{table}

\noindent \textbf{Edge Case Subset:} During our experiments, we observed that in many cases, the friction location on an object did not significantly affect its future position and orientation during short pushing actions. However, we hypothesise that friction will have a more pronounced effect in longer interactive tasks. In particular, edge cases, where the object is pushed for the full 4 seconds, present scenarios where the friction location creates outcomes that are difficult to predict using vision alone. These edge cases are crucial for exploring our problem, so we created a subset of four test cases, illustrated in Fig.~\ref{fig:EdgedDataSeq}. In these test cases, the object starts from the same location, with friction applied at each corner of the box, respectively. Despite identical robot actions, the final positions of the object differ drastically, making these cases ideal for qualitative analysis.

\chadded[id=AE]{While four representative trials are illustrated for qualitative comparison in Fig.~\ref{fig:EdgedDataSeq}, quantitative metrics reported in Table 3 are computed over the full edge-case test subset (550 unseen friction configurations and 50 seen configurations), ensuring statistical robustness.}

\section{Results, Evaluation and Analysis}
\label{results}
\subsection{\chadded[id=AE]{When Does Tactile--Visual Integration Help?}}
\label{sec:spots_help}
\chadded[id=AE]{Before presenting quantitative results, it is important to clarify the conditions under which tactile--visual integration is expected to improve predictive performance. When object dynamics are visually distinguishable and physical properties can be inferred from appearance alone, vision-based models often perform competitively. In such cases, tactile input provides limited additional information, and only marginal performance gains are expected.}

\chadded[id=AE]{In contrast, when physical properties such as friction, contact location, or compliance are visually ambiguous or entirely invisible, tactile feedback becomes critical. In these scenarios, prediction accuracy depends on the model’s ability to integrate cross-modal signals and update its internal representation during interaction. The experiments in this section are explicitly designed to separate these two regimes.}

\chreplaced[id=AE]{In this section, we investigate whether integrating tactile and visual sensations during physical robot interactions can enhance an agent's understanding of cause-and-effect relationships. To this end, we compare (i) the same video prediction system, SVG, with and without tactile sensation, and (ii) the same tactile prediction system, ACTP, with and without visual sensation.}{In this section, we investigate whether and under what conditions integrating tactile and visual sensations during physical robot interactions enhances an agent’s understanding of physical cause-and-effect relationships. To this end, we compare (i) a vision-based video prediction system (SVG) with and without tactile input, and (ii) a tactile prediction system (ACTP) with and without visual input.}

The tactile- and vision-integrated versions of SVG and ACTP are described in Section~\ref{section:method}, and all tested model variants are summarised in Table~\ref{tab:keyfeatures}. Using these models, we conduct a comparative study that isolates the architectural and sensory contributions of each prediction pipeline.
\chreplaced[id=AE]{The primary objective of this research is to determine whether the integration of tactile and visual modalities can improve an agent's predictive capabilities.} {The primary objective of this study is to determine when the integration of tactile and visual modalities improves an agent’s predictive capabilities beyond unimodal baselines.}

To assess this, we evaluate the proposed models on two action-conditioned tactile pushing datasets. Specifically, our experiments are designed to: (i) evaluate scene prediction performance relative to a vision-only baseline (SVG); (ii) assess generalisation to unseen objects; (iii) compare alternative multi-modal prediction architectures; (iv) examine the role of tactile input during inference via sensory removal; (v) evaluate tactile prediction performance relative to a tactile-only baseline (ACTP); and (vi) analyse the dependence of tactile prediction on visual input through systematic visual occlusion.

\textit{Scene Evaluation Metrics.} We evaluate scene prediction performance using Peak Signal-to-Noise Ratio (PSNR), Structural Similarity Index (SSIM), and Mean Absolute Error (MAE), which provide complementary pixel-wise measures of visual fidelity. Although the marked object dataset could enable trajectory-based metrics using the marker centroids, this approach is not applicable here because predicted scenes do not explicitly reconstruct these markers. In addition to quantitative evaluation, we perform qualitative analysis with a particular focus on physically ambiguous edge cases, where visual appearance alone is insufficient to explain interaction outcomes.

\textit{Tactile Evaluation Metrics.} Following prior work on tactile prediction with magnetic-based sensors \cite{mandil2021tac}, we use Mean Absolute Error (MAE) as the primary quantitative metric. However, as established in earlier studies, numerical errors alone provide limited insight into tactile prediction quality. Accordingly, we complement quantitative evaluation with qualitative analysis of individual taxel force trajectories over extended time horizons, focusing on the accurate prediction of force transients, peaks, and changes in magnitude. As with scene evaluation, particular emphasis is placed on edge case interactions that expose differences between competing models.

\textit{Training and Test Procedure.} Scene images are resized to $\mathbf{v}_i \in \mathbb{R}^{64 \times 64 \times 3}$. All models are trained end-to-end using the stochastic video prediction framework described in \cite{denton2018stochastic}, with the learned prior replacing the inference network at test time. Implementations are realised in PyTorch, and training and evaluation are performed on two NVIDIA RTX A6000 GPUs.

%%%%%%%%%%%%%%%%%%%%%%%%%%%%%%%%%%%
%%%%%%%%%%%%%%%%%%%%%%%%%%%%%%%%%%%

\begin{table}[t]
    \centering
    \begin{tabular}{llll}
        \hline
        \textbf{Model}       & \textbf{MAE $\downarrow$} & \textbf{PSNR $\uparrow$} & \textbf{SSIM  $\uparrow$} \\
        \hline
        SVG & 0.0100 $\pm$ \scriptsize{4.8$e^{-4}$} & 81.1243 $\pm$ \scriptsize{2.8$e^{-1}$} & 0.9809 $\pm$ \scriptsize{1.4$e^{-3}$} \\
        SVG-TE & 0.0100 $\pm$ \scriptsize{9.3$e^{-4}$} & 81.1274 $\pm$ \scriptsize{5.4$e^{-1}$} & 0.9808 $\pm$ \scriptsize{2.9$e^{-3}$} \\
        SVTG & 0.0109 $\pm$ \scriptsize{3.2$e^{-4}$} & 80.3639 $\pm$ \scriptsize{1.7$e^{-1}$} & 0.9783 $\pm$ \scriptsize{1.0$e^{-3}$} \\
        SPOTS & \textbf{0.0099} $\pm$ \scriptsize{4.3$e^{-4}$} & 81.1979 $\pm$ \scriptsize{2.5$e^{-1}$} & 0.9812 $\pm$ \scriptsize{1.3$e^{-3}$} \\
        SPOTS-small & 0.0099 $\pm$ \scriptsize{1.7$e^{-3}$} & \textbf{81.2247} $\pm$ \scriptsize{9.8$e^{-1}$} & \textbf{0.9812} $\pm$ \scriptsize{5.1$e^{-3}$} \\  
        SPOTS-SOP & 0.0100 $\pm$ \scriptsize{4.5$e^{-4}$} & 81.1424 $\pm$ \scriptsize{2.6$e^{-1}$} & 0.9809 $\pm$ \scriptsize{1.3$e^{-3}$} \\
        \\
               & \textbf{MAE i+5 $\downarrow$} & \textbf{PSNR i+5 $\uparrow$} & \textbf{SSIM i+5 $\uparrow$} \\
        \hline   
        SVG & 0.0112 $\pm$ \scriptsize{5.9$e^{-5}$} & 79.5207 $\pm$ \scriptsize{3.3$e^{-2}$} & 0.9766 $\pm$ \scriptsize{1.8$e^{-4}$} \\
        SVG-TE & 0.0112 $\pm$ \scriptsize{2.2$e^{-5}$} & 79.5484 $\pm$ \scriptsize{2.2$e^{-2}$} & \textbf{0.9767} $\pm$ \scriptsize{8.9$e^{-5}$} \\
        SVTG & 0.0129 $\pm$ \scriptsize{5.5$e^{-5}$} & 78.6714 $\pm$ \scriptsize{1.7$e^{-2}$} & 0.9715 $\pm$ \scriptsize{2.0$e^{-4}$} \\
        SPOTS & 0.0113 $\pm$ \scriptsize{4.8$e^{-5}$} & 79.5417 $\pm$ \scriptsize{1.5$e^{-2}$} & 0.9766 $\pm$ \scriptsize{1.4$e^{-4}$} \\
        SPOTS-small & \textbf{0.0112} $\pm$ \scriptsize{5.0$e^{-5}$} & \textbf{79.5703} $\pm$ \scriptsize{3.7$e^{-2}$} & 0.9767 $\pm$ \scriptsize{1.1$e^{-4}$} \\
        SPOTS-SOP & 0.0114 $\pm$ \scriptsize{7.8$e^{-5}$} & 79.5014 $\pm$ \scriptsize{3.5$e^{-2}$} & 0.9764 $\pm$ \scriptsize{2.5$e^{-4}$} \\
        \hline
    \end{tabular}
    \caption{\textit{Household object clusters dataset}: Average scene prediction performance on both the combined seen and unseen household object cluster test datasets. Prediction scores are presented with 95\% confidence intervals. The i+5 scores represent the metrics at the 5th prediction frame in the prediction horizon, which is the maximum used during training.}
    \label{tab:ScePrePerformance_HC}
\end{table}
%%%%%%%%%%%%%%%%%%%%%%%%%%%%%%%%%%%
%%%%%%%%%%%%%%%%%%%%%%%%%%%%%%%%%%%

\begin{figure}[t]
    \centering
    \adjustbox{fbox=0pt 4pt, frame}{\includegraphics[trim=0pt 0pt 0pt 4pt, clip, width=0.99\textwidth]{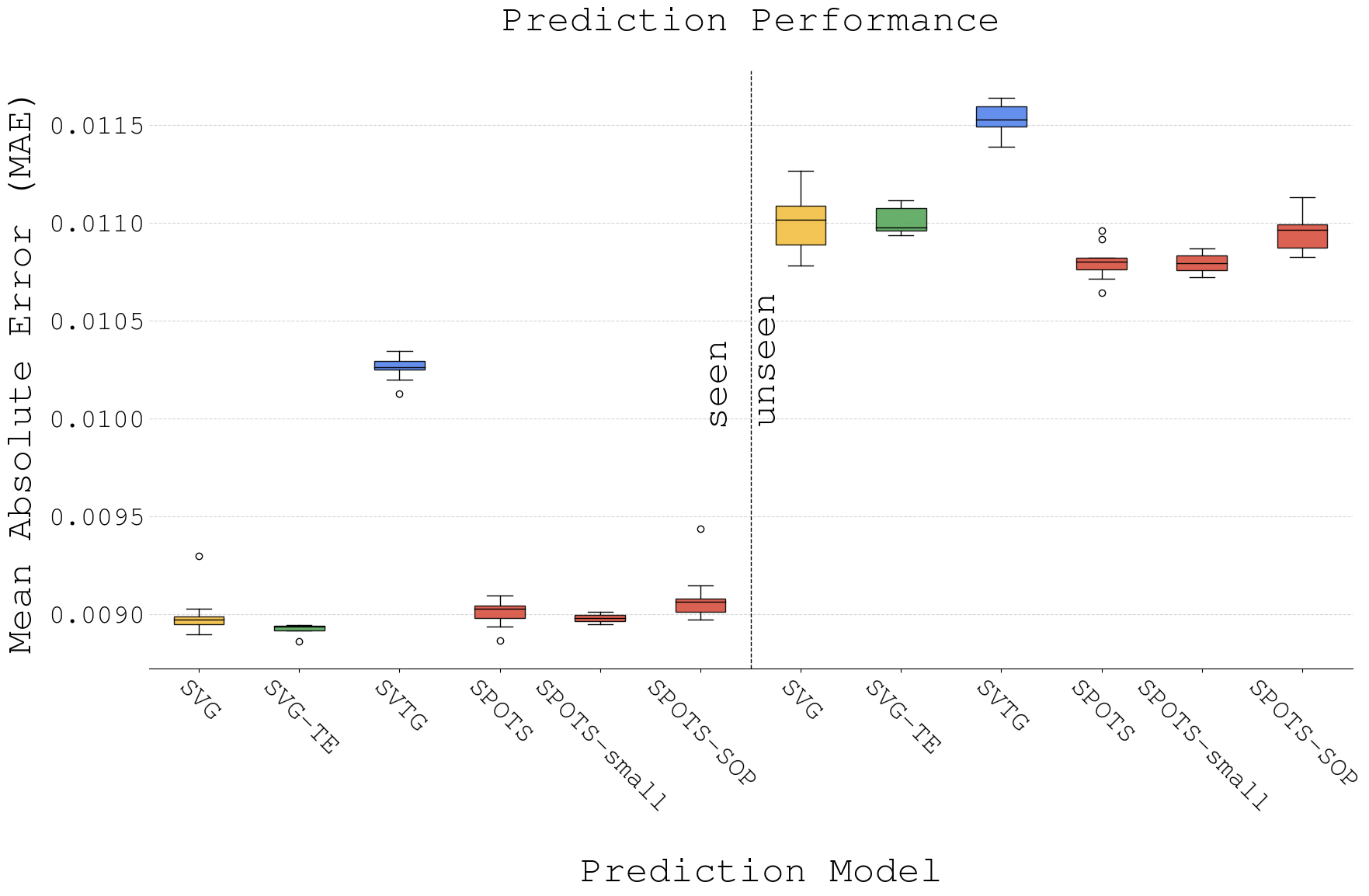}}
    \caption{\textit{Household object clusters dataset}: The Mean Absolute Error (MAE) performance metric for models on seen (left) and unseen (right) object clusters. Each model was trained 10 times, and its performance statistics are shown in these box and whisker plots. SVG is without tactile sensation. SVG-TE does not predict tactile sensation. SVTG and SPOTS models predict both touch and vision, using single and dual-pipeline architectures, respectively. SPOTS result in the best performances for unseen objects.}
    \label{fig:SSIM_HC_performance}
\end{figure}

%%%%%%%%%%%%%%%%%%%%%%%%%%%%%%%%%%%%%%%%%%%%%%%%%%%%%%%%%%%%%%%%%%%%%%%%%%%%%%%%%%%%%%%%%%%%

\begin{table*}[t!]
    \centering
    \resizebox{\textwidth}{!}{%
    \begin{tabular}{lllll}
        \hline
        Model       & MAE $\downarrow$  & MAE i+5 $\downarrow$  & PSNR $\uparrow$ & SSIM $\uparrow$ \\
        \hline
        \multicolumn{5}{l}{\textbf{Visually Identical Dataset}}  \\
        \hline
        SVG                 & 0.0100 $\pm$ \scriptsize{9.73$\times 10^{-5}$} & 0.0122 $\pm$ \scriptsize{1.59$\times 10^{-4}$} & 78.7258 $\pm$ \scriptsize{5.65$\times 10^{-2}$} & 0.9647 $\pm$ \scriptsize{3.52$\times 10^{-4}$} \\
        SVG-TE              & \textbf{0.0098} $\pm$ \scriptsize{9.37$\times 10^{-5}$} & 0.0120 $\pm$ \scriptsize{1.96$\times 10^{-4}$} & 78.9524 $\pm$ \scriptsize{6.61$\times 10^{-2}$} & 0.9659 $\pm$ \scriptsize{4.05$\times 10^{-4}$} \\
        SVTG                & 0.0109 $\pm$ \scriptsize{1.33$\times 10^{-4}$} & 0.0129 $\pm$ \scriptsize{3.33$\times 10^{-4}$} & 79.0638 $\pm$ \scriptsize{6.99$\times 10^{-2}$} & 0.9630 $\pm$ \scriptsize{3.14$\times 10^{-4}$} \\
        SPOTS                & 0.0099 $\pm$ \scriptsize{1.89$\times 10^{-4}$} & 0.0119 $\pm$ \scriptsize{2.39$\times 10^{-4}$} & 79.0778 $\pm$ \scriptsize{1.20$\times 10^{-1}$} & 0.9661 $\pm$ \scriptsize{8.34$\times 10^{-4}$} \\
        SPOTS-small          & 0.0099 $\pm$ \scriptsize{1.39$\times 10^{-4}$} & 0.0120 $\pm$ \scriptsize{2.19$\times 10^{-4}$} & 79.0938 $\pm$ \scriptsize{1.32$\times 10^{-1}$} & 0.9660 $\pm$ \scriptsize{7.93$\times 10^{-4}$} \\
        SPOTS-SOP            & 0.0099 $\pm$ \scriptsize{6.82$\times 10^{-5}$} & \textbf{0.0119} $\pm$ \scriptsize{1.37$\times 10^{-4}$} & \textbf{79.1263} $\pm$ \scriptsize{8.09$\times 10^{-2}$} & \textbf{0.9662} $\pm$ \scriptsize{5.50$\times 10^{-4}$} \\
        \hline
        \multicolumn{5}{l}{\textbf{Visually Identical Edge Case Subset}}  \\
        \hline
        SVG                 & 0.0104 $\pm$ \scriptsize{1.19$\times 10^{-4}$} & 0.0129 $\pm$ \scriptsize{4.71$\times 10^{-4}$} & 77.5338 $\pm$ \scriptsize{1.18$\times 10^{-2}$} & 0.9586 $\pm$ \scriptsize{9.95$\times 10^{-4}$} \\
        SVG-TE              & 0.0095 $\pm$ \scriptsize{2.34$\times 10^{-4}$} & 0.0116 $\pm$ \scriptsize{4.10$\times 10^{-4}$} & 78.6620 $\pm$ \scriptsize{2.69$\times 10^{-1}$} & 0.9650 $\pm$ \scriptsize{1.62$\times 10^{-3}$} \\
        SVTG                & 0.0101 $\pm$ \scriptsize{2.30$\times 10^{-4}$} & 0.0115 $\pm$ \scriptsize{3.04$\times 10^{-4}$} & 78.8688 $\pm$ \scriptsize{4.18$\times 10^{-1}$} & 0.9627 $\pm$ \scriptsize{2.20$\times 10^{-3}$} \\
        SPOTS                & 0.0093 $\pm$ \scriptsize{3.33$\times 10^{-4}$} & 0.0113 $\pm$ \scriptsize{5.49$\times 10^{-4}$} & 79.1245 $\pm$ \scriptsize{2.32$\times 10^{-1}$} & 0.9666 $\pm$ \scriptsize{1.24$\times 10^{-3}$} \\
        SPOTS-small          & \textbf{0.0091} $\pm$ \scriptsize{2.30$\times 10^{-4}$} & \textbf{0.0110} $\pm$ \scriptsize{2.11$\times 10^{-4}$} & \textbf{79.2577} $\pm$ \scriptsize{2.38$\times 10^{-1}$} & \textbf{0.9671} $\pm$ \scriptsize{1.07$\times 10^{-3}$} \\
        SPOTS-SOP            & 0.0092 $\pm$ \scriptsize{1.66$\times 10^{-4}$} & 0.0111 $\pm$ \scriptsize{3.11$\times 10^{-4}$} & 79.1381 $\pm$ \scriptsize{1.54$\times 10^{-1}$} & 0.9666 $\pm$ \scriptsize{5.37$\times 10^{-4}$} \\
        \hline
        \multicolumn{5}{l}{\textbf{Visually Identical Edge Case Subset - Anaesthetised}}  \\
        \hline
        SVG                 & \textbf{0.0104} $\pm$ \scriptsize{1.19$\times 10^{-4}$} & \textbf{0.0129} $\pm$ \scriptsize{4.71$\times 10^{-4}$} & \textbf{77.5335} $\pm$ \scriptsize{1.18$\times 10^{-2}$} & \textbf{0.9586} $\pm$ \scriptsize{9.95$\times 10^{-4}$} \\
        SVG-TE              & 0.0113 $\pm$ \scriptsize{6.34$\times 10^{-4}$} & 0.0145 $\pm$ \scriptsize{1.13$\times 10^{-3}$} & 76.8562 $\pm$ \scriptsize{5.58$\times 10^{-1}$} & 0.9542 $\pm$ \scriptsize{4.10$\times 10^{-3}$} \\
        SVTG                & 0.0217 $\pm$ \scriptsize{4.69$\times 10^{-3}$} & 0.0271 $\pm$ \scriptsize{9.01$\times 10^{-3}$} & 72.4685 $\pm$ \scriptsize{1.80$\times 10^{0}$} & 0.9014 $\pm$ \scriptsize{2.12$\times 10^{-2}$} \\
        SPOTS                & 0.0108 $\pm$ \scriptsize{2.86$\times 10^{-4}$} & 0.0131 $\pm$ \scriptsize{6.85$\times 10^{-4}$} & 77.5330 $\pm$ \scriptsize{2.59$\times 10^{-1}$} & 0.9575 $\pm$ \scriptsize{1.73$\times 10^{-3}$} \\
        SPOTS-small          & 0.0113 $\pm$ \scriptsize{6.27$\times 10^{-4}$} & 0.0135 $\pm$ \scriptsize{8.44$\times 10^{-4}$} & 77.1867 $\pm$ \scriptsize{4.48$\times 10^{-1}$} & 0.9549 $\pm$ \scriptsize{3.60$\times 10^{-3}$} \\
        SPOTS-SOP            & 0.0110 $\pm$ \scriptsize{8.38$\times 10^{-4}$} & 0.0132 $\pm$ \scriptsize{1.04$\times 10^{-3}$} & 77.4204 $\pm$ \scriptsize{6.48$\times 10^{-1}$} & 0.9568 $\pm$ \scriptsize{4.59$\times 10^{-3}$} \\
        \hline
    \end{tabular}}
    \caption{\textit{Visually identical dataset, the edge case subset, anaesthetised test case}: Average scene prediction performance on the , including 95\% confidence intervals.}
    \label{tab:ScePrePerformance_MO}
\end{table*}
%%%%%%%%%%%%%%%%%%%%%%%%%%%%%%%%%%%%%%%%%%%%%%%%%%%%%%%%%%%%%%%%%%%%%%%%%%%%%%%%%%%%%%%%%%%%

\begin{figure}[t]
    \centering
    \adjustbox{fbox=0pt 4pt, frame}{\includegraphics[trim=0pt 0pt 0pt 0pt, clip, height=0.37\textwidth]{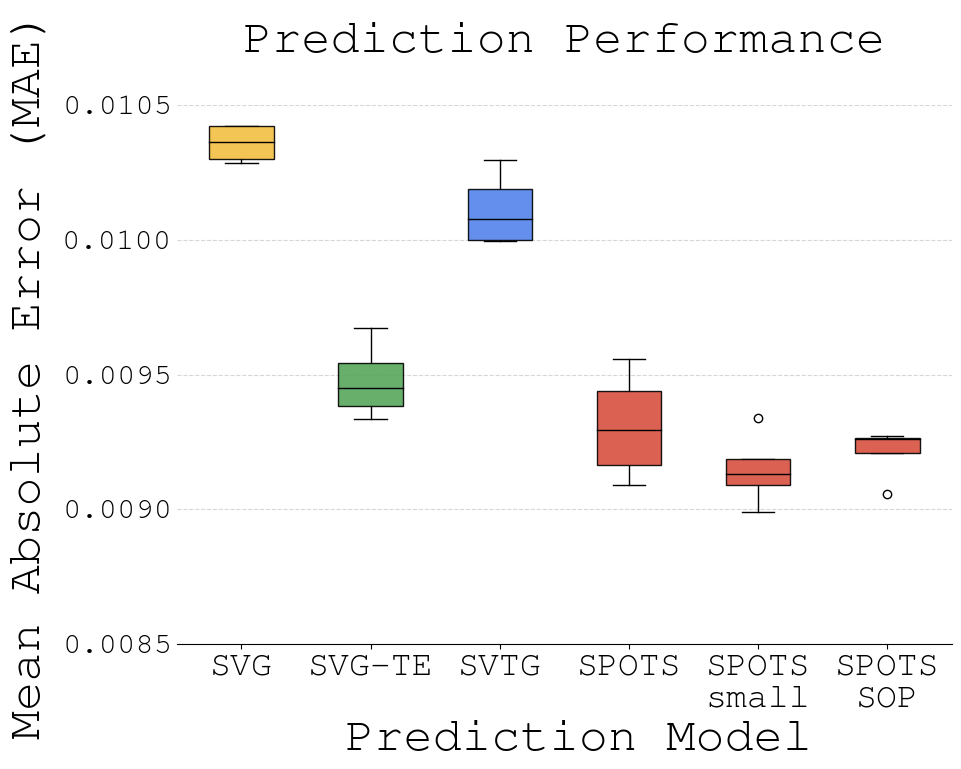}}
    \adjustbox{fbox=0pt 4pt, frame}{\includegraphics[trim=0pt 0pt 0pt 0pt, clip, height=0.37\textwidth]{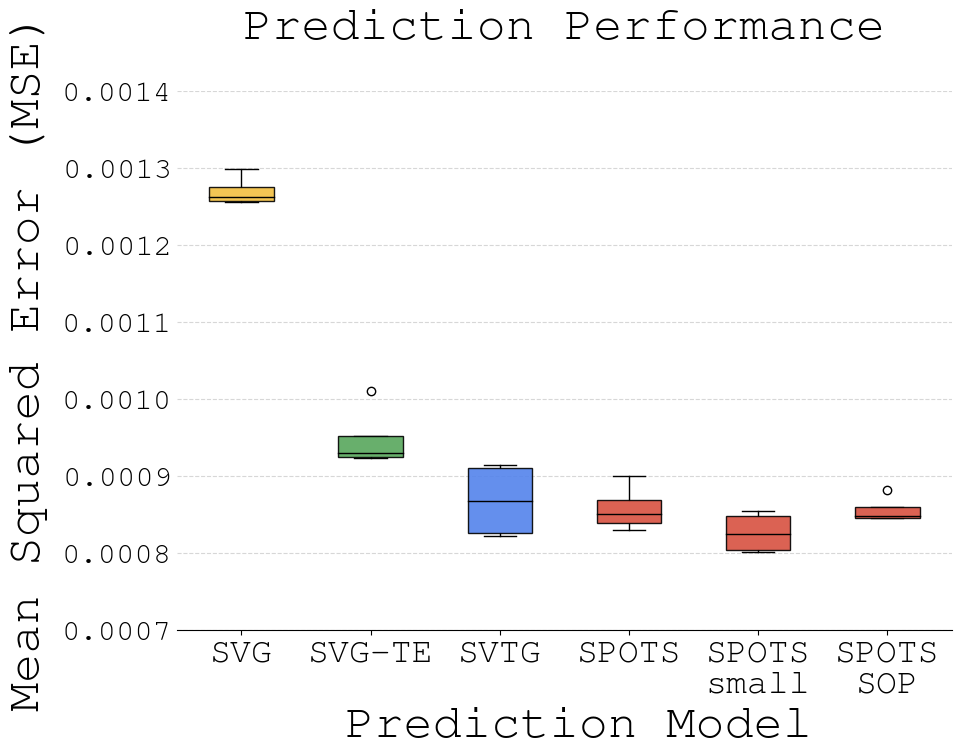}}
    \caption{\textit{Visually Identical Edge Case Subset}: The Mean Absolute Error (MAE) and Mean Squared Error (MSE) performance for prediction models. Each model was trained 10 times to generate these performance statistics. SVG is without tactile sensation. SVG-TE does not predict tactile sensation. SVTG and SPOTS models predict both touch and vision, using single and dual-pipeline architectures, respectively. The tactile-enabled models outperform the vision-only model, with SPOTS-small achieving the best performance overall, which is consistent with the regime distinction discussed in Section~6.1.}
    \label{fig:SSIM_MO_performance}
\end{figure}

\subsection{\chreplaced[id=AE]{Performance on Visually Unambiguous Object Clusters}
{Quantitative Scene Analysis}}

\chreplaced[id=AE]{Table~\ref{tab:ScePrePerformance_HC} reports average scene prediction performance on the household object cluster dataset, which primarily contains visually distinguishable objects. As discussed in Section~\ref{sec:spots_help}, improvements are not expected in visually
unambiguous interaction regimes.}{Table~\ref{tab:ScePrePerformance_HC} reports average scene prediction performance on the Household Cluster dataset, which predominantly contains visually distinguishable objects. Across this dataset, all models achieve similar performance, and performance differences between vision-only and tactile-enabled models remain modest. This behaviour is expected, as the underlying physical properties of many objects can be inferred directly from visual appearance.}
\chreplaced[id=AE]{Importantly, SPOTS does not degrade performance despite its dual-pipeline structure and achieves comparable or slightly improved generalisation on unseen object clusters.}{Importantly, the proposed dual-pipeline architecture does not degrade visual prediction performance. SPOTS and SPOTS-small achieve performance comparable to, and in some cases slightly better than, vision-only baselines on unseen object clusters, indicating that the architecture preserves strong visual modelling capacity while remaining robust to additional sensory input.}

In the household cluster data set, small deviations in test performance suggest minimal differences between models enabled by tactile devices and the baseline only with vision. Notably, SPOTS-small demonstrates the best overall performance across both the full prediction horizon and the final time step, despite having a parameter count comparable to SVG.
The breakdown of model performance is shown in Fig.~\ref{fig:SSIM_HC_performance}, where the results are separated into groups that contain seen and unseen objects. For objects observed during training, SVG, SVG-TE, and SPOTS models perform similarly. However, for clusters of unseen objects, the SPOTS-based models consistently outperform SVG and SVG-TE. This indicates that while both tactile-enabled and vision-only models can accurately predict interactions involving familiar objects, SPOTS models generalize more effectively to previously unseen object instances.

\chadded[id=AE]{Although several confidence intervals overlap, performance trends remain consistent across independent training runs, indicating stable architectural behaviour rather than variance-driven effects.}

\chreplaced[id=AE]{ The results in Table \ref{tab:ScePrePerformance_MO} across the Visually identical test dataset and the Edge Case subset verify the findings from the Household Cluster dataset.}{ The results reported in Table~\ref{tab:ScePrePerformance_MO} for the Visually Identical dataset and its Edge Case subset extend these findings to physically ambiguous scenarios, and are analysed in detail in the following subsection.}
\chdeleted[id=AE]{ Multi-modal models generally outperform the vision-only model (except SVTG). The significant performance improvements on the edge case subset highlight that tactile-enabled models can better handle scenarios where visually identical scenes have different physical properties. Interestingly, the parameter size differences between tactile-enabled models do not seem to impact performance significantly, as SPOTS-small (with a similar number of weights as SVG) is still effective in these settings. Furthermore, the smaller SPOTS model even outperforms its larger counterpart, suggesting that a smaller model size might be beneficial.}
\chdeleted[id=AE]{ The single-pipeline multi-modal prediction system SVTG performs worse than SVG across the entire household dataset, but excels on the visually identical dataset. This indicates that while SVTG may produce blurred visual predictions in scenes where tactile sensation is unnecessary, it effectively utilises tactile information in scenes where it is essential. Hence, the poorer visual prediction performance reflects a higher degree of object blur rather than a lack of understanding of physical interactions. }
\chdeleted[id=AE]{ The box-and-whisker plots in Fig. \ref{fig:SSIM_MO_performance} emphasise the superiority of dual-pipeline systems (SPOTS, SPOTS-STP, and SPOTS-small) over other scene prediction methods, highlighting their enhanced grasp of physical dynamics. }
\chadded[id=AE]{Our proposed dual-pipeline models maintain strong visual prediction performance and exhibit improved generalisation to unseen objects without incurring a performance penalty. These findings provide a stable reference point against which the benefits of tactile-visual integration under physical ambiguity are assessed in the following sections.}

%%%%%%%%%%%%%%
%%%%%%%%%%%%%%
%%%%%%%%%%%%%%

\subsection{\chreplaced[id=AE]{Performance under Physical Ambiguity}{Qualitative Scene Analysis}}
\label{QualitativeSceneAnalysis}

\chadded[id=AE]{The visually identical object dataset (Table~\ref{tab:ScePrePerformance_MO}) isolates interaction scenarios in which visual appearance alone is insufficient to infer underlying physical properties. In this setting, accurate prediction relies on effective integration of tactile feedback during interaction, as reflected by the lower average error of SPOTS-small across repeated trials on the visually identical edge-case subset. Figure~\ref{fig:SSIM_MO_performance} quantifies this effect on the visually identical edge-case subset,}  \chreplaced[id=AE]{showing that tactile-enabled models tend to achieve lower MAE and MSE than the vision-only baseline, with SPOTS-small achieving the lowest average error across runs.}{showing that tactile-enabled models consistently achieve lower MAE and MSE than the vision-only baseline, with SPOTS-small performing best overall.}

Quantitative analysis provides general insights into prediction performance, focusing on pixel-wise comparisons across the entire image. However, to assess an agent's perception of physical interaction, qualitative analysis is crucial. The key aspect evaluated here is the location of the interacted object, as the precise location is essential for understanding physical interactions, even if other factors like object crispness are also important in some applications.
\chadded[id=AE]{Accordingly, the following qualitative analysis focuses on object location accuracy in physically ambiguous interactions, using edge-case examples to highlight differences that are not captured by pixel-wise metrics alone (Figure~\ref{fig:edge_cases_test} and \ref{fig:HC_test_2}).}

\begin{figure}[tb!]
    \centering
    \adjustbox{fbox=0pt 4pt, frame}{\includegraphics[trim=1cm 0pt 0pt 0pt, clip, height=0.474\textwidth]{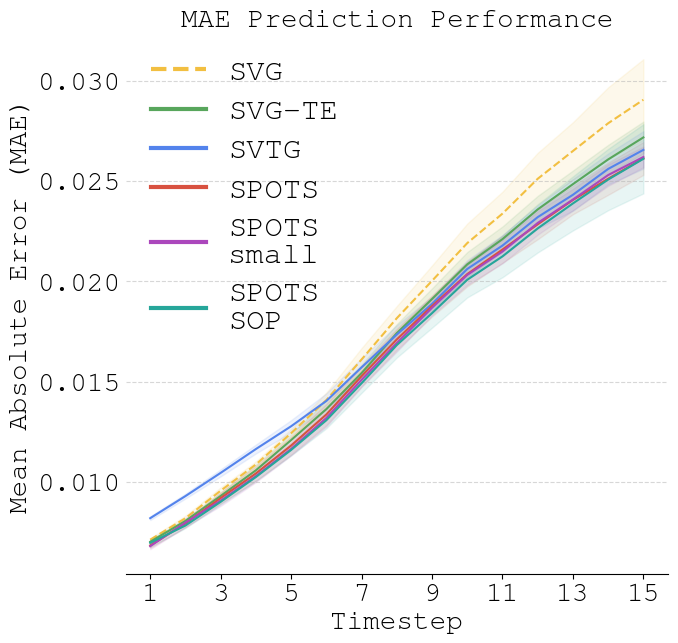}}
    \adjustbox{fbox=0pt 4pt, frame}{\includegraphics[trim=1cm 0pt 0pt 0pt, clip, height=0.474\textwidth]{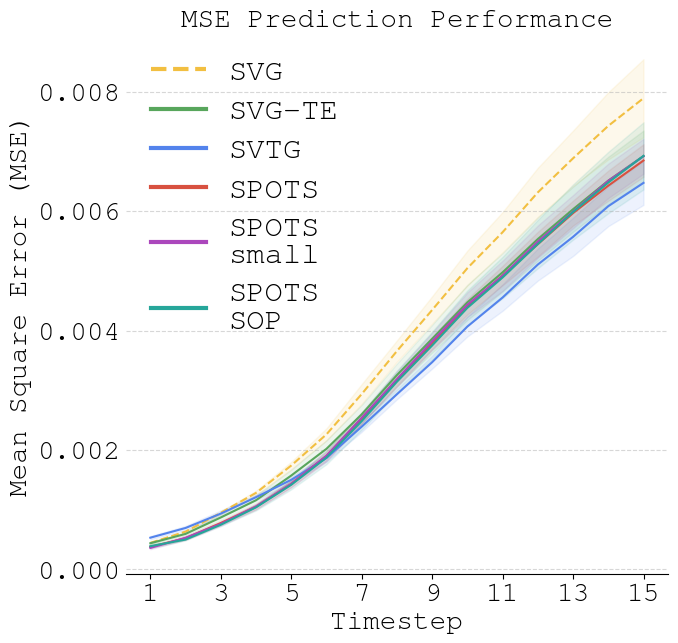}}
    % \adjustbox{fbox=0pt 4pt, frame}{\includegraphics[trim=1cm 0pt 0pt 0pt, clip, height=0.3\textwidth]{Results/multi-models/Google_style/MHO_QUANT_LTH_SSIM_LSTM.png}}
    \caption{\textit{Visually Identical Edge Case Subset}: These diagrams illustrate the prediction performance over an extended time series horizon (15 prediction frames). The models were trained for up to 5 prediction frames. MAE and MSE performance metrics reveal that over longer prediction horizons, the tactile-enabled models progressively outperform the non-tactile model SVG.}
    \label{fig:SSIM_and_MAE_loostimeseries}
\end{figure}

\subsection{\chadded[id=AE]{Long-Horizon Prediction and Error Accumulation}}

\chadded[id=AE]{While short-horizon predictions show limited differences between models, error accumulation over longer horizons reveals a widening performance gap. Tactile-enabled models, particularly SPOTS, degrade more gracefully as uncertainty increases, indicating improved internal modelling of physical dynamics.} 

Fig.~\ref{fig:SSIM_and_MAE_loostimeseries} illustrates prediction performance over extended horizons beyond the training window. As prediction uncertainty accumulates, tactile-enabled models increasingly outperform the vision-only baseline SVG. Although SVG performs comparably during early time steps, its limited access to physical interaction cues leads to compounding errors at longer horizons. In contrast, tactile-enabled models leverage contact information to update their internal state, resulting in more stable and accurate long-term predictions.

\begin{figure*}[t!]
    \centering
    \adjustbox{fbox=0pt 4pt, frame}{\includegraphics[height=0.35\textwidth]{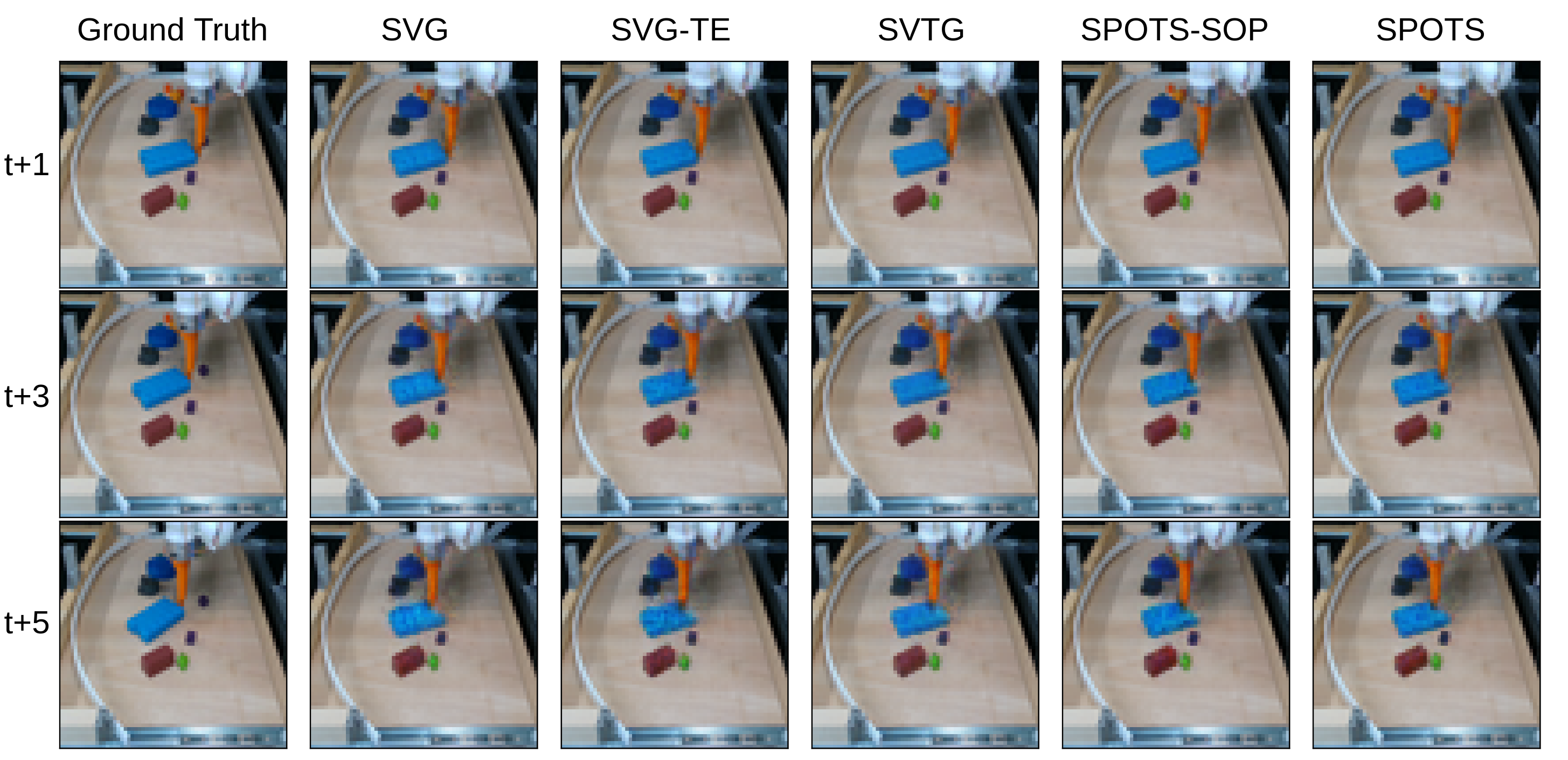}}
    \caption{\textit{House Cluster Dataset}: Comparison of different prediction models on the household cluster test set for time-steps $\{i+1, i+3, i+5\}$ into the prediction horizon. The models shown are the highest validation-scoring models from their respective batches. We observe little qualitative difference between the models on this dataset.}
    \label{fig:edge_cases_test}
\end{figure*}

\chreplaced[id=AE]{Likewise, while the single-pipeline SVTG model underperforms in early time steps,its improved modelling of cause-effect dynamics enables it to maintain competitive performance relative to the vision-only SVG model over extended periods.} {Similarly, although the single-pipeline SVTG model underperforms at short horizons due to increased visual blur, it surpasses the vision-only SVG baseline at longer horizons, suggesting that access to tactile information improves its representation of underlying physical dynamics despite weaker visual fidelity.}
\chdeleted[id=AE]{ Fig. \ref{fig:edge_cases_test} shows model performance on the household cluster dataset, where minimal deviation in predictions makes qualitative analysis challenging. }
\chreplaced[id=AE]{ Nevertheless, the visually identical dataset and the edge case datasets offer deeper insights into the models' physical interaction perception.} {To further interpret these long-horizon trends, qualitative results on physically ambiguous interactions provide additional insight into model behaviour.}

\begin{figure*}[tb!]
    \centering
    \adjustbox{fbox=0pt 4pt, frame}{\includegraphics[width=0.8\textwidth]{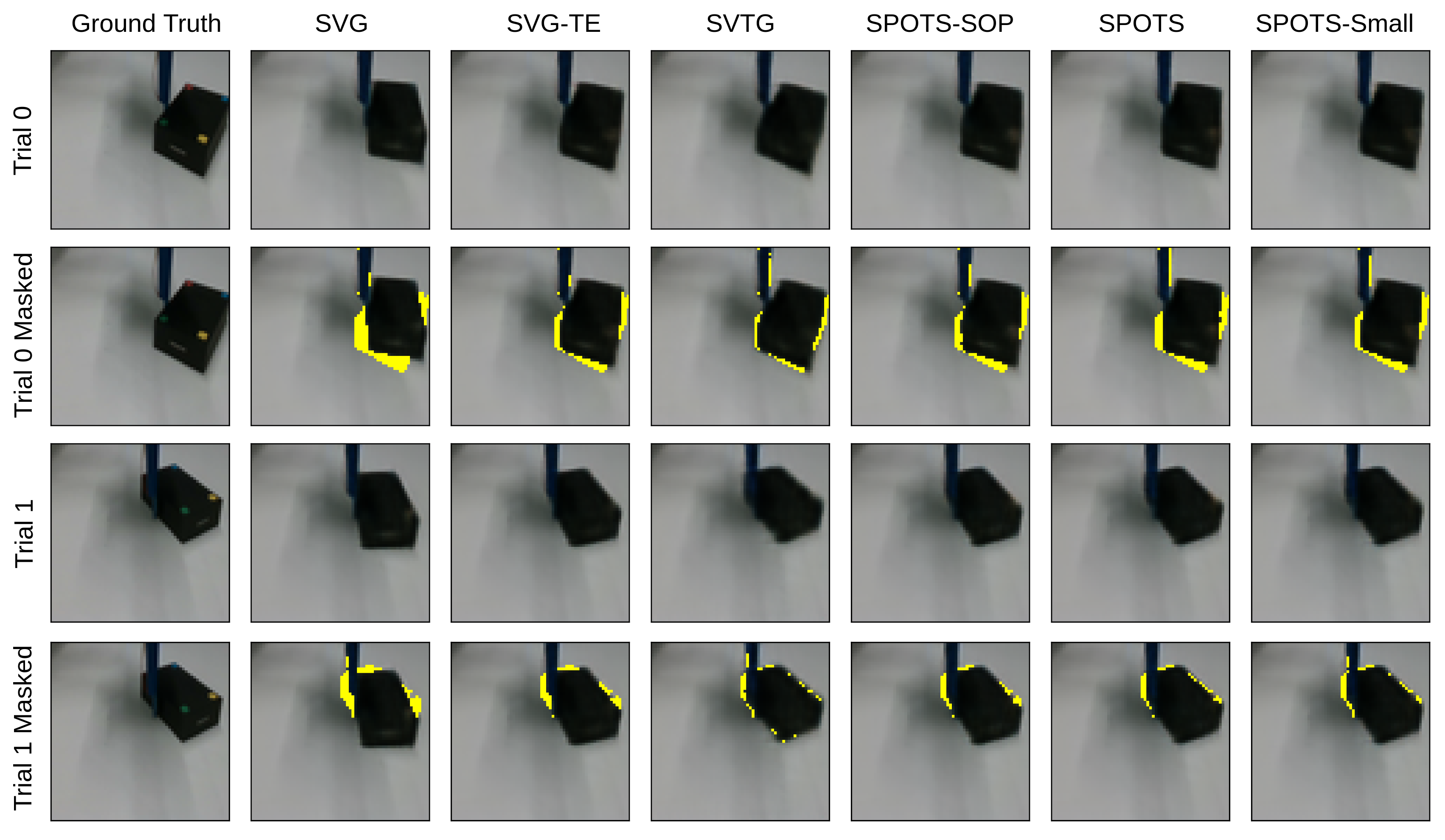}}
    \caption{\textit{Visually Identical Edge Case Subset}: Comparison of different prediction models on the edge case test subset shown in Figure \ref{fig:EdgedDataSeq}. The models depicted are the highest-scoring ones from the 10 trained models. Predictions are displayed for timestep i+5 across the different models. The masked rows show the ground truth object locations in yellow with the predicted object image overlaid.}
    \label{fig:HC_test_2}
\end{figure*}

Fig.~\ref{fig:HC_test_2} presents prediction results at time step $i+5$ for edge case trials shown in Fig.~\ref{fig:EdgedDataSeq}. Rows 2 and 4 overlay the ground truth object location in yellow behind the predicted object position, highlighting spatial prediction errors. Smaller yellow regions indicate more accurate object localisation.

\chreplaced[id=AE]{ The tactile-enabled models generally predict object locations more accurately. While the SVG model struggles with physical property understanding, the SVTG model, despite lower quantitative scores, predicts the most accurate object locations, indicating superior physical interaction perception.} {Tactile-enabled models generally predict object locations more accurately than the vision-only baseline. While SVG struggles to resolve latent physical properties, both SPOTS and SVTG capture contact-driven object motion more faithfully, indicating improved physical interaction perception.}

\chreplaced[id=AE]{ In alignment with quantitative analysis, model size has little impact on physical interaction perception, with both SPOTS and SPOTS-small delivering similarly strong predictions.} {Consistent with quantitative results, model size has a limited impact on long-horizon physical interaction perception, with SPOTS and SPOTS-small exhibiting comparable qualitative behaviour.}
\chreplaced[id=AE]{ Overall, tactile sensation consistently improves video prediction trends across ambiguous scenarios. Although differences remain modest in absolute magnitude and confidence intervals partially overlap, tactile-enabled models exhibit more stable long-horizon behaviour. Despite SPOTS’ stronger quantitative scores, SVTG shows slightly better qualitative localisation performance. Taken together, these results suggest that tactile input provides meaningful predictive structure under physical ambiguity.} {Overall, these results demonstrate that tactile input plays a crucial role in mitigating error accumulation during long-horizon prediction. While architectural differences influence visual sharpness and short-term accuracy, access to tactile feedback consistently improves the model’s ability to maintain accurate object motion estimates over time, particularly in physically ambiguous interactions.}

%%%%%%%%%%%%%%%%%%%%%%%%%%%%%%%%%%%
%%%%%%%%%%%%%%%%%%%%%%%%%%%%%%%%%%%
%%%%%%%%%%%%%%%%%%%%%%%%%%%%%%%%%%%
%%%%%%%%%%%%%%%%%%%%%%%%%%%%%%%%%%%

%%%%%%%%%%%%%%
%%%%%%%%%%%%%%
%%%%%%%%%%%%%%

%%%%%%%%%%%%%%%%%%%%%%%%%%%%%%%%%%%%%%%%%%
\begin{figure*}[tb!]
    \centering
    \adjustbox{fbox=0pt 4pt, frame}{\includegraphics[width=0.8\textwidth]{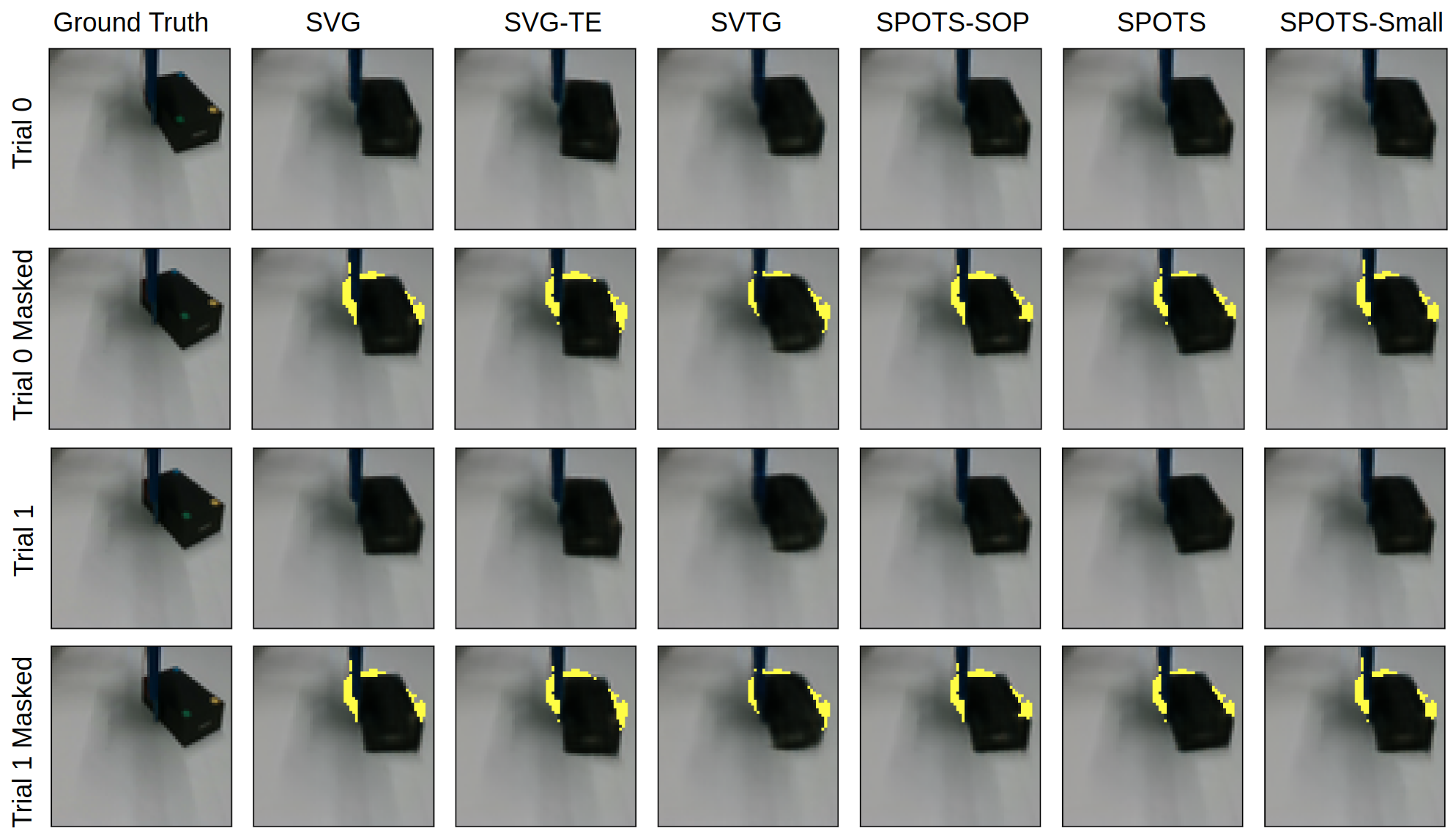}}
    \caption{Comparison of different prediction models using mean tactile signal values (anaesthetisation) on the edge case test subset shown in Figure \ref{fig:EdgedDataSeq}. Predictions are shown for time-step $i+5$. The results suggest that while training with tactile data improves model performance, the models still rely heavily on live tactile input for accurate predictions.}
    \label{fig:EC_test_no_tactile}
\end{figure*}
%%%%%%%%%%%%%%%%%%%%%%%%%%%%%%%%%%%%%%%%%%

\subsection{\chreplaced[id=AE]{Ablation via Sensory Removal}
{Anaesthetisation of Prediction Models}}

\chadded[id=AE]{To verify that performance improvements arise from active cross-modal integration rather than increased model capacity, we evaluate the models under systematic sensory removal, including tactile anaesthetisation and visual occlusion.}

To further evaluate the impact of tactile sensation on physical interaction perception, we test each model by replacing tactile data with values corresponding to no contact. By "anaesthetising" the agent's fingers, we effectively remove tactile input, allowing us to assess its importance. Figure \ref{fig:EC_test_no_tactile} shows the qualitative results of this test, and quantitative results are presented in Table \ref{tab:ScePrePerformance_MO}. Both sets of results reveal that the prediction performance of the tactile-enabled models is equal to the non-tactile-enabled model, SVG. SVTG trades visual fidelity for improved physical localisation, explaining its lower pixel-level metrics but stronger qualitative performance.

This experiment demonstrates the importance of tactile sensation for an agent's cause-and-effect understanding, enabling the multi-modal models to update their internal understanding of the system in real-time. Without tactile input, tactile-enabled models revert to performance levels comparable to non-tactile models, indicating that the learning process itself does not significantly enhance visual perception, but that the tactile modality as an input does.

%%%%%%%%%%%%%%
%%%%%%%%%%%%%%
%%%%%%%%%%%%%%

\subsection{\chreplaced[id=AE]{Tactile Prediction Performance}{Tactile Prediction Analysis}}
%\chadded[id=AE]{To verify that performance improvements arise from active cross-modal integration rather than increased model capacity, we evaluate the models under systematic sensory removal, including tactile anaesthetisation and visual occlusion.}

%%%%%%%%%%%%%%%%%%%%%%%%%%%%%%%%%%%%%%%%%%%%%%%
\begin{figure*}[t!]
    \centering
    \subfloat[]{\adjustbox{fbox=0pt 0pt, frame}{\includegraphics[trim=0pt 120pt 0pt 0pt, clip, width=0.32\textwidth]{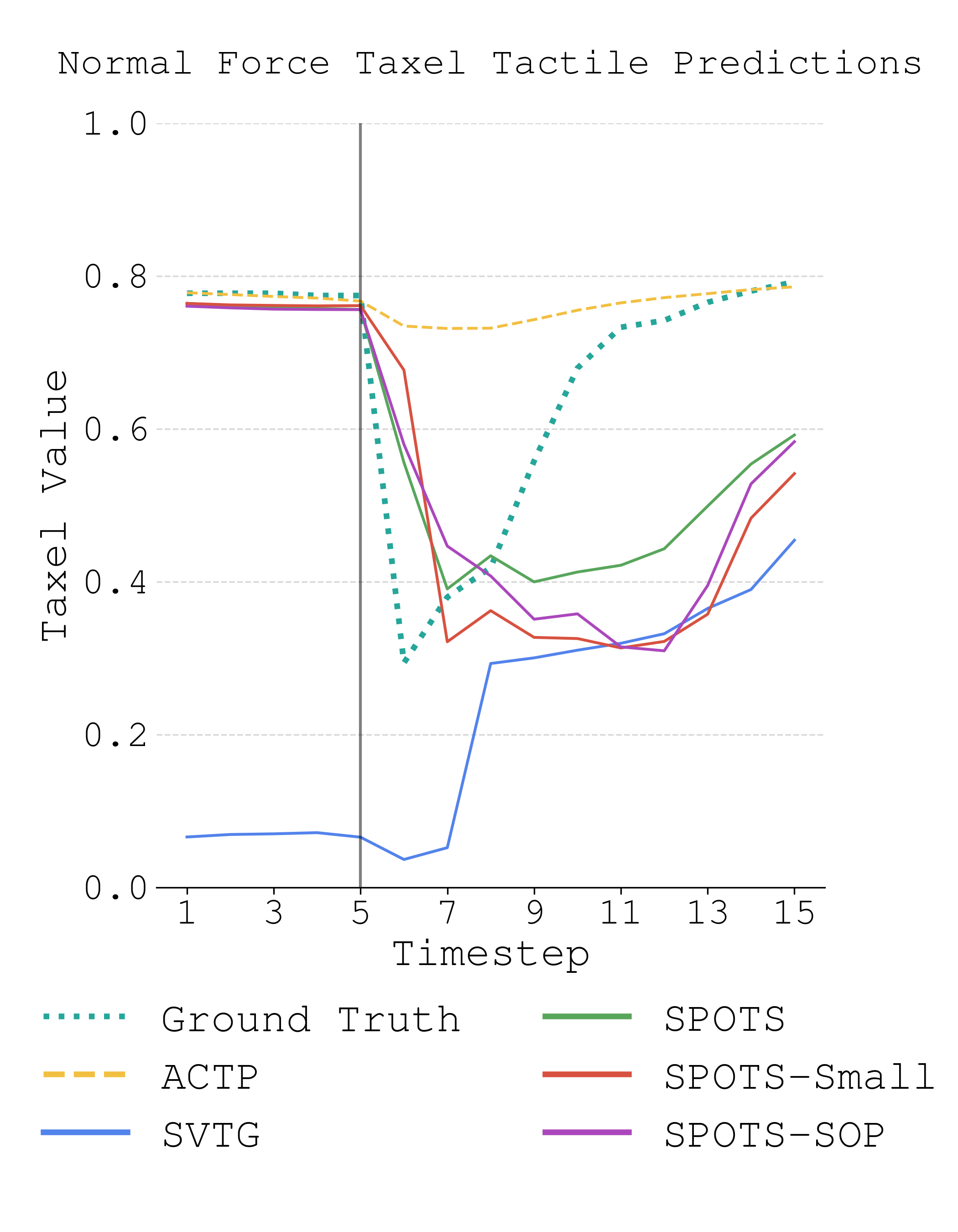}}}
    \hspace{0.1cm}
    \subfloat[]{\adjustbox{fbox=0pt 0pt, frame}{\includegraphics[trim=0pt 120pt 0pt 0pt, clip, width=0.32\textwidth]{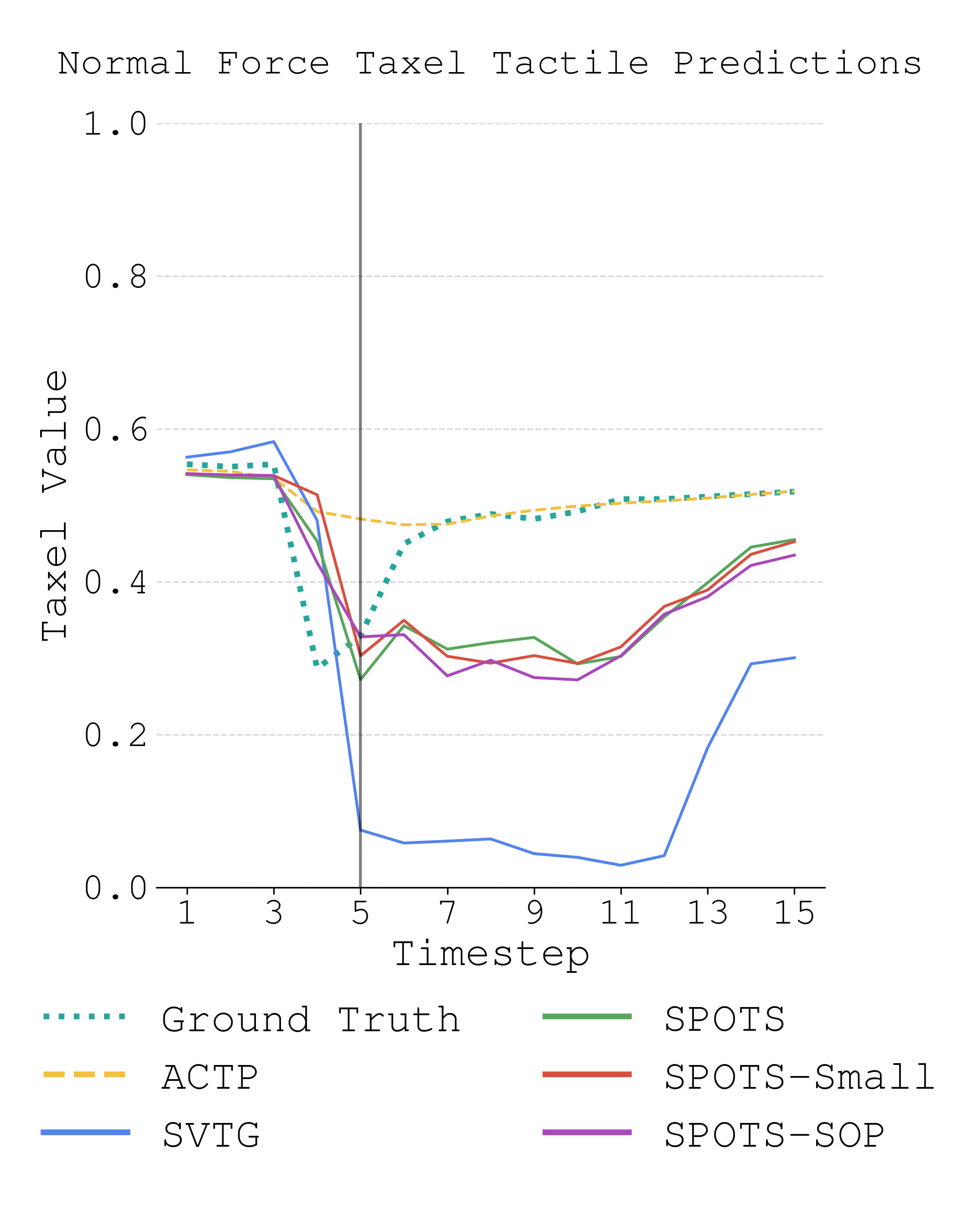}}}
    \hspace{0.1cm}
    \subfloat[]{\adjustbox{fbox=0pt 0pt, frame}{\includegraphics[trim=0pt 120pt 0pt 0pt, clip, width=0.32\textwidth]{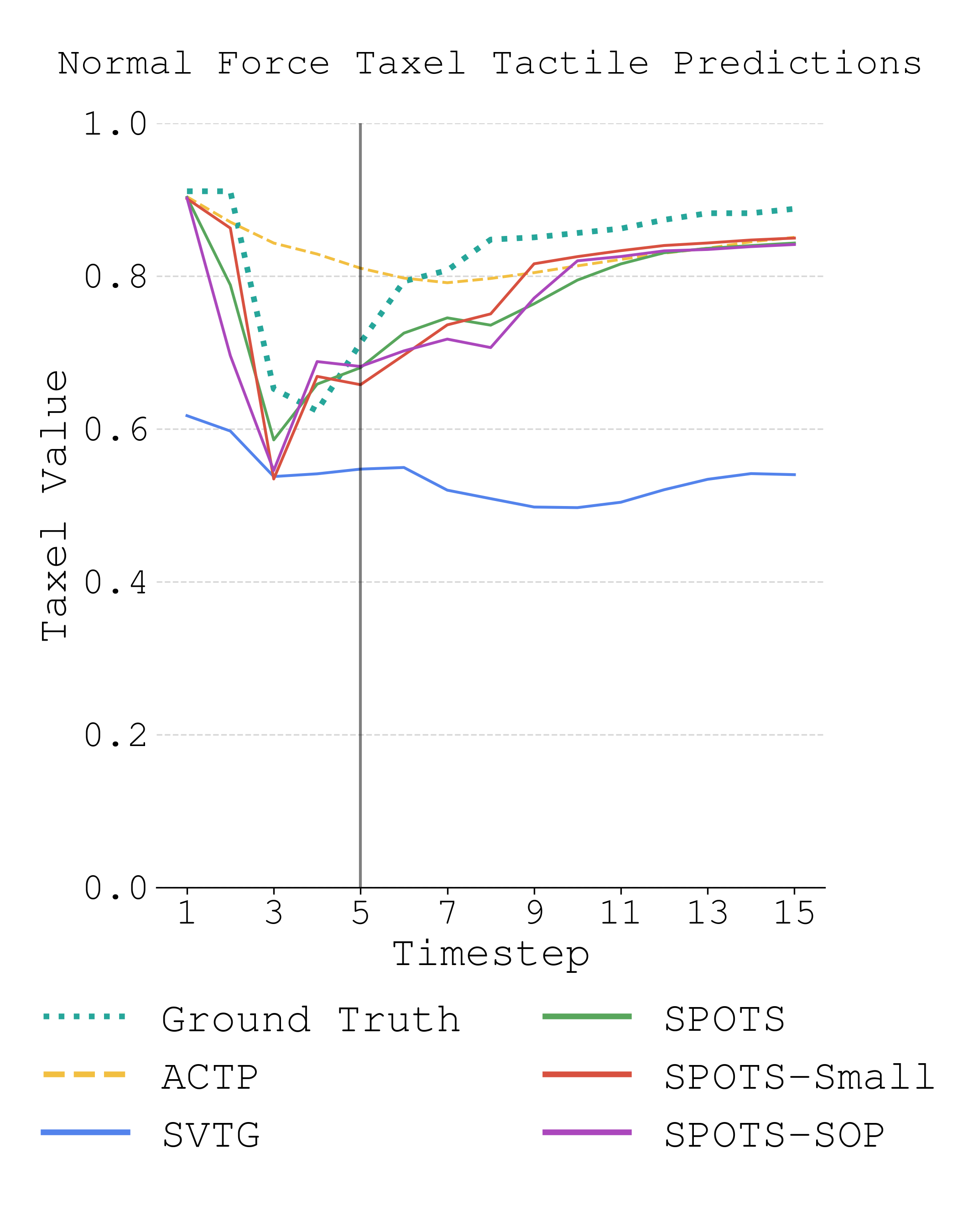}}}

    \subfloat[]{\adjustbox{fbox=0pt 0pt, frame}{\includegraphics[trim=0pt 20pt 0pt 0pt, clip, width=0.32\textwidth]{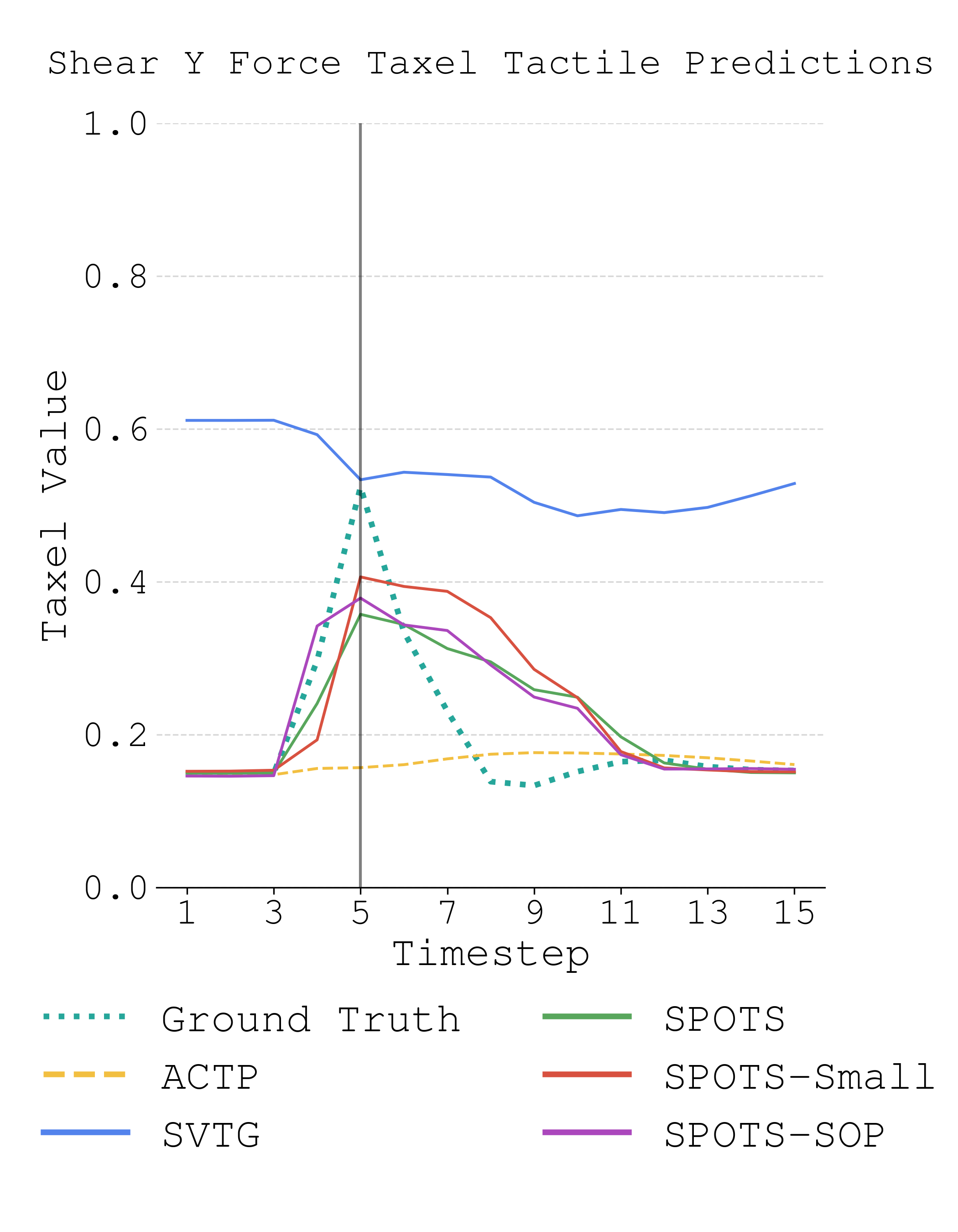}}}
    \hspace{0.1cm}
    \subfloat[]{\adjustbox{fbox=0pt 0pt, frame}{\includegraphics[trim=0pt 20pt 0pt 0pt, clip, width=0.32\textwidth]{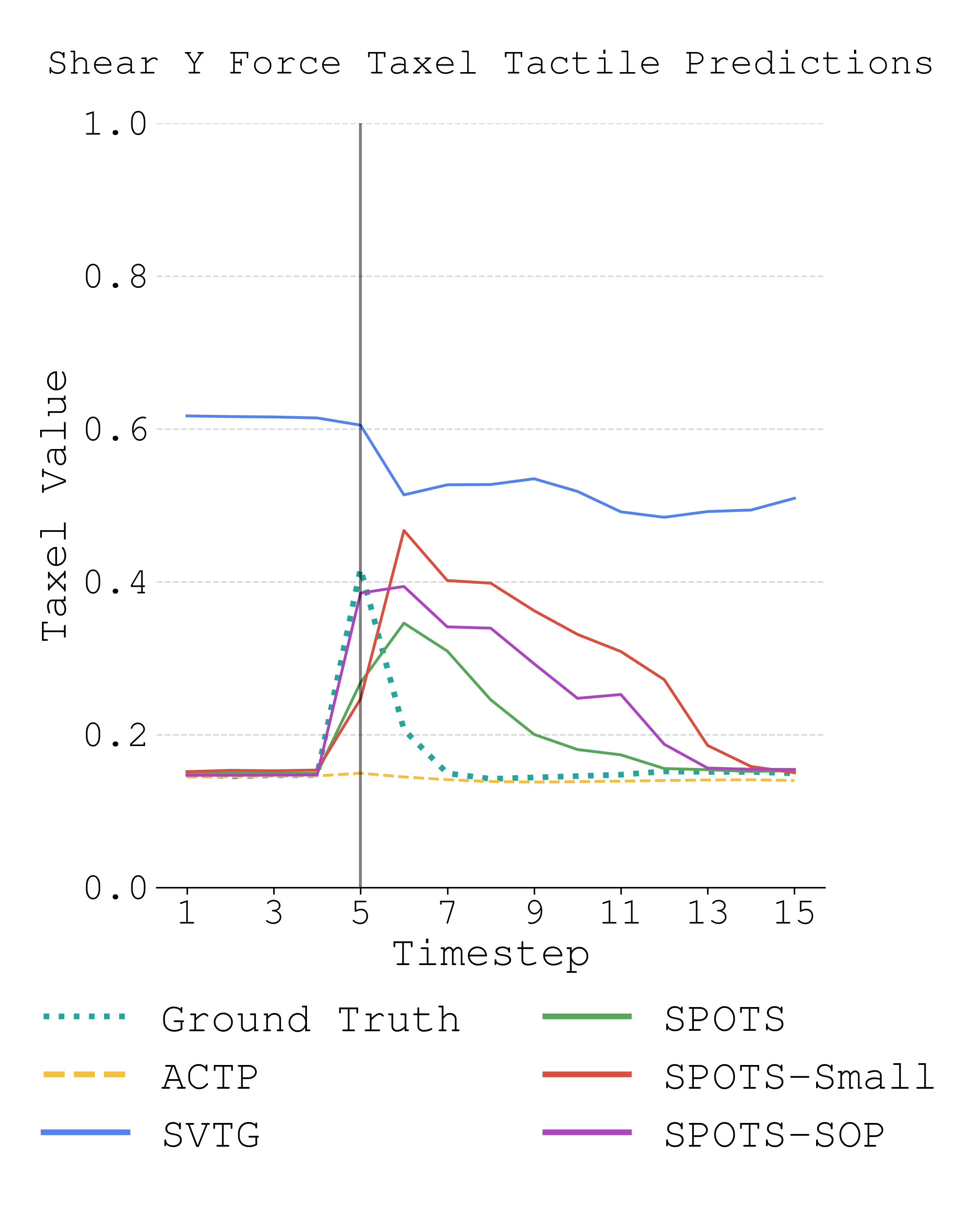}}}
    \hspace{0.1cm}
    \subfloat[]{\adjustbox{fbox=0pt 0pt, frame}{\includegraphics[trim=0pt 20pt 0pt 0pt, clip, width=0.32\textwidth]{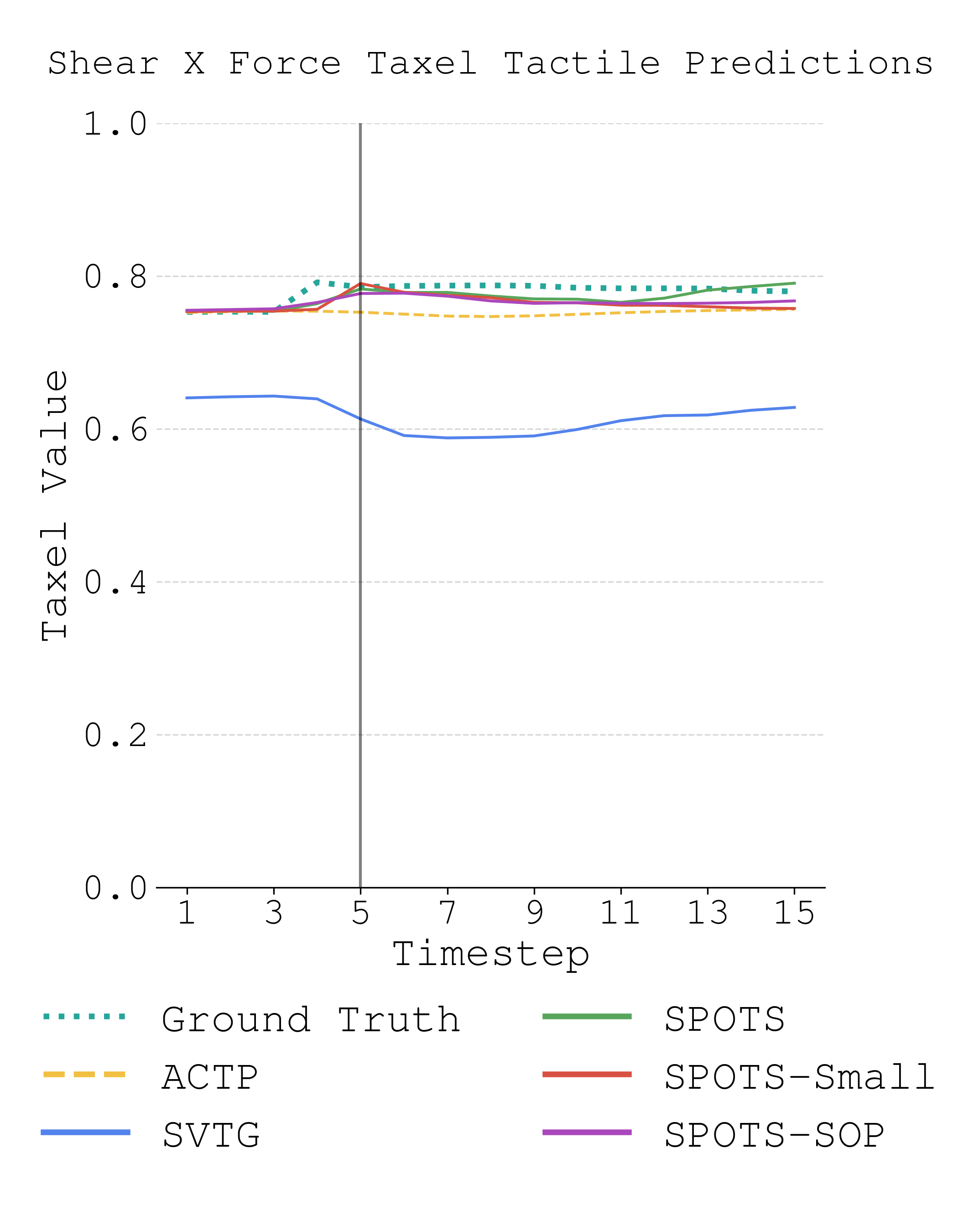}}}
    \caption{\textit{Visually Identical Edge Case Subset}: Tactile predictions of a single force from a single magnetic sensing element for six separate cases. Each graph shows a single Normal Force (a-c), Shear Y (d-e), or Shear X (f) taxel from the center of the tactile sensor. SPOTS models outperform the state-of-the-art tactile-only prediction model ACTP.}
    \label{fig:Tactile_Prediction}
\end{figure*}
%%%%%%%%%%%%%%%%%%%%%%%%%%%%%%%%%%%%%%%%%%

%%%%%%%%%%%%%%%%%%%%%%%%%
%%%%%%%%%%%%%%%%%%%%%%%%%
\begin{table}[tb!]
    \centering
    \begin{tabular}{llll}
        \hline
        Model       & MAE $\downarrow$ & MAE $i+5$ $\downarrow$ & MAE $i+15$ $\downarrow$ \\
        \hline
        ACTP        & 0.0452 $\pm$ \scriptsize{0.0124} & 0.0379 $\pm$ \scriptsize{0.0024} & 0.0488 $\pm$ \scriptsize{0.0052} \\
        SVTG        & 0.2584 $\pm$ \scriptsize{0.0006} & 0.2664 $\pm$ \scriptsize{0.0005} & 0.2452 $\pm$ \scriptsize{0.0010} \\
        SPOTS       & 0.0167 $\pm$ \scriptsize{0.0056} & \textbf{0.0116} $\pm$ \scriptsize{0.0003} & 0.0261 $\pm$ \scriptsize{0.0017} \\
        SPOTS-small & 0.0167 $\pm$ \scriptsize{0.0063} & \textbf{0.0116} $\pm$ \scriptsize{0.0003} & 0.0262 $\pm$ \scriptsize{0.0005} \\
        SPOTS-SOP   & \textbf{0.0165} $\pm$ \scriptsize{0.0055} & 0.0118 $\pm$ \scriptsize{0.0005} & \textbf{0.0261} $\pm$ \scriptsize{0.0008} \\
        \hline
    \end{tabular}
    \caption{Average tactile prediction performance on edge case marked object test dataset. Mean Absolute Error (MAE) scores are shown for the extended prediction horizon of 15 time steps, including time steps 5 and 15. Confidence intervals (95\%) are included.}
    \label{tab:TacPredPerformance}
\end{table}
%%%%%%%%%%%%%%%%%%%%%%%%%
%%%%%%%%%%%%%%%%%%%%%%%%%
Table \ref{tab:TacPredPerformance} presents the quantitative performance of each model, measured by Mean Absolute Error (MAE) across the entire prediction horizon, as well as for specific time steps. Figure \ref{fig:Tactile_Prediction} shows the qualitative performance of each model on the test set. 

The SPOTS models outperform the the tactile only prediction model ACTP and the single pipeline prediction model SVTG. The SVTG model uses SVG style architecture to predict magnetic-based tactile sensations, which has produced equally poor results in \cite{mandil2021tac}. The dual pipeline prediction architectures (SPOTS) are able to perform accurate predictions in both the tactile and scene prediction spaces by utilising the architectures that work best for the given modality. In this case, the single pipeline model SVTG, which worked well for visual predictions, was not capable of producing the same for tactile predictions as its rigid architecture could not be adapted to fit both modalities. 

Qualitative analysis shown in Fig. \ref{fig:Tactile_Prediction} shows the poor performance of ACTP, which shows very little change in force predictions over the horizon. In line with the poor MAE scores, SVTG predicts poor tactile sensations that often produce little to no resemblance to the ground truth tactile values. The dual pipeline (SPOTS) models produce good estimates of changes in tactile sensation, predicting spikes at the right time steps and with roughly correct scale. These SPOTS predictions dramatically outperform the state-of-the-art tactile prediction model ACTP and the single pipeline model SVTG. Post spike the tactile predictions can linger at high/low values, for example in Fig \ref{fig:Tactile_Prediction} (a). There is little to distinguish the three SPOTS models from each other, so we conclude that the dual pipeline structure is the main cause for tactile prediction success and the small changes in the SPOTS architecture have a negligible impact. 

\chdeleted[id=AE]{Overall, we find that, both quantitatively and qualitatively, the SPOTS model is capable of generating realistic tactile predictions across long prediction horizons. In the next subsection, we will analyse the impact that visual sensation has on the tactile prediction aspect of our models, is it only useful for training or does the model leverage live scene data to improve its tactile predictions. }

%%%%%%%%%%%%%%
%%%%%%%%%%%%%%
%%%%%%%%%%%%%%

\begin{figure*}[t]
    \centering
    \subfloat[]{\adjustbox{fbox=0pt 4pt, frame}{\includegraphics[trim=20pt 45pt 0pt 20pt, clip, width=0.3\textwidth]{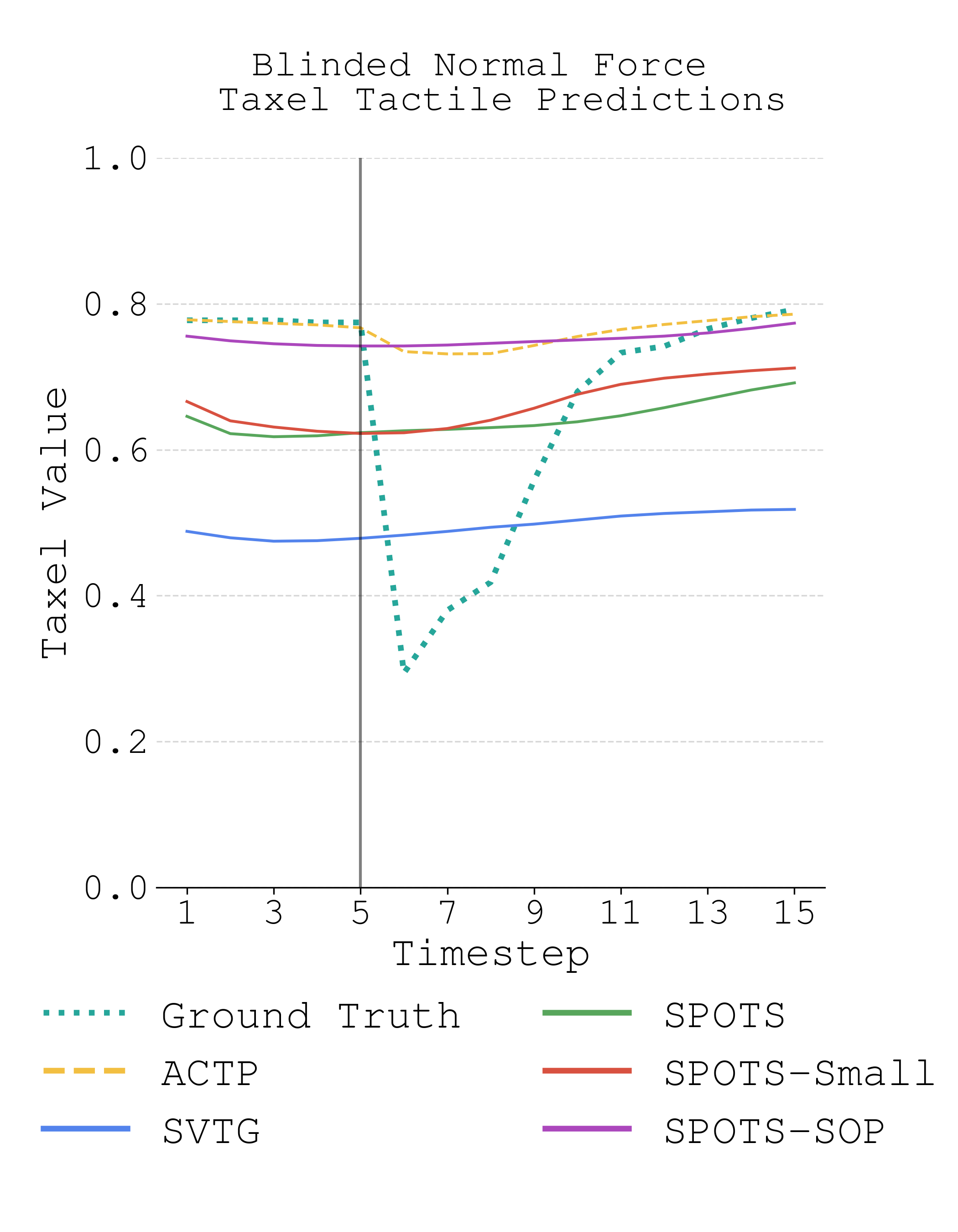}}}
    \hspace{0.1cm}
    \subfloat[]{\adjustbox{fbox=0pt 4pt, frame}{\includegraphics[trim=20pt 45pt 0pt 20pt, clip, width=0.3\textwidth]{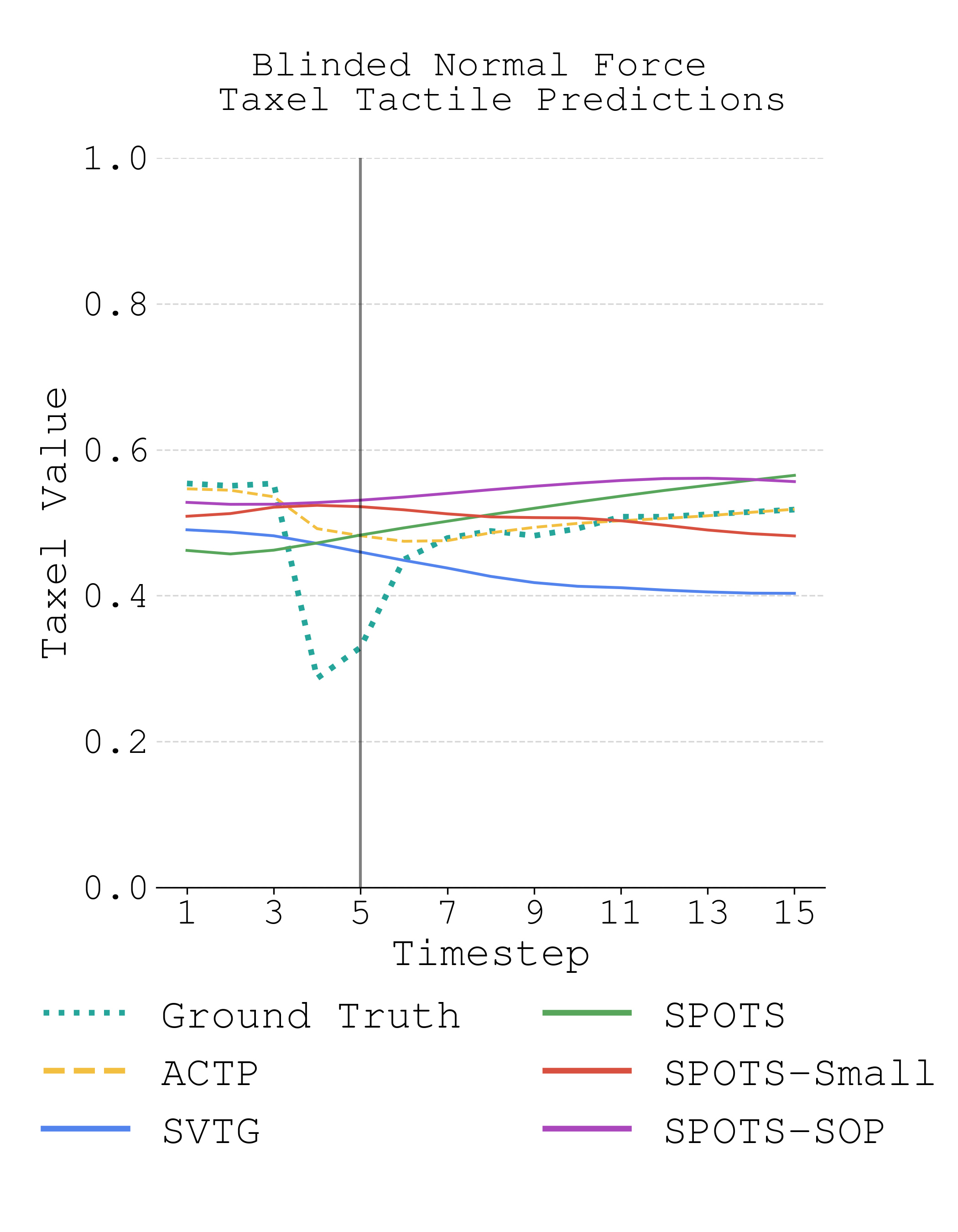}}}
    \hspace{0.1cm}
    \subfloat[]{\adjustbox{fbox=0pt 4pt, frame}{\includegraphics[trim=20pt 45pt 0pt 20pt, clip, width=0.3\textwidth]{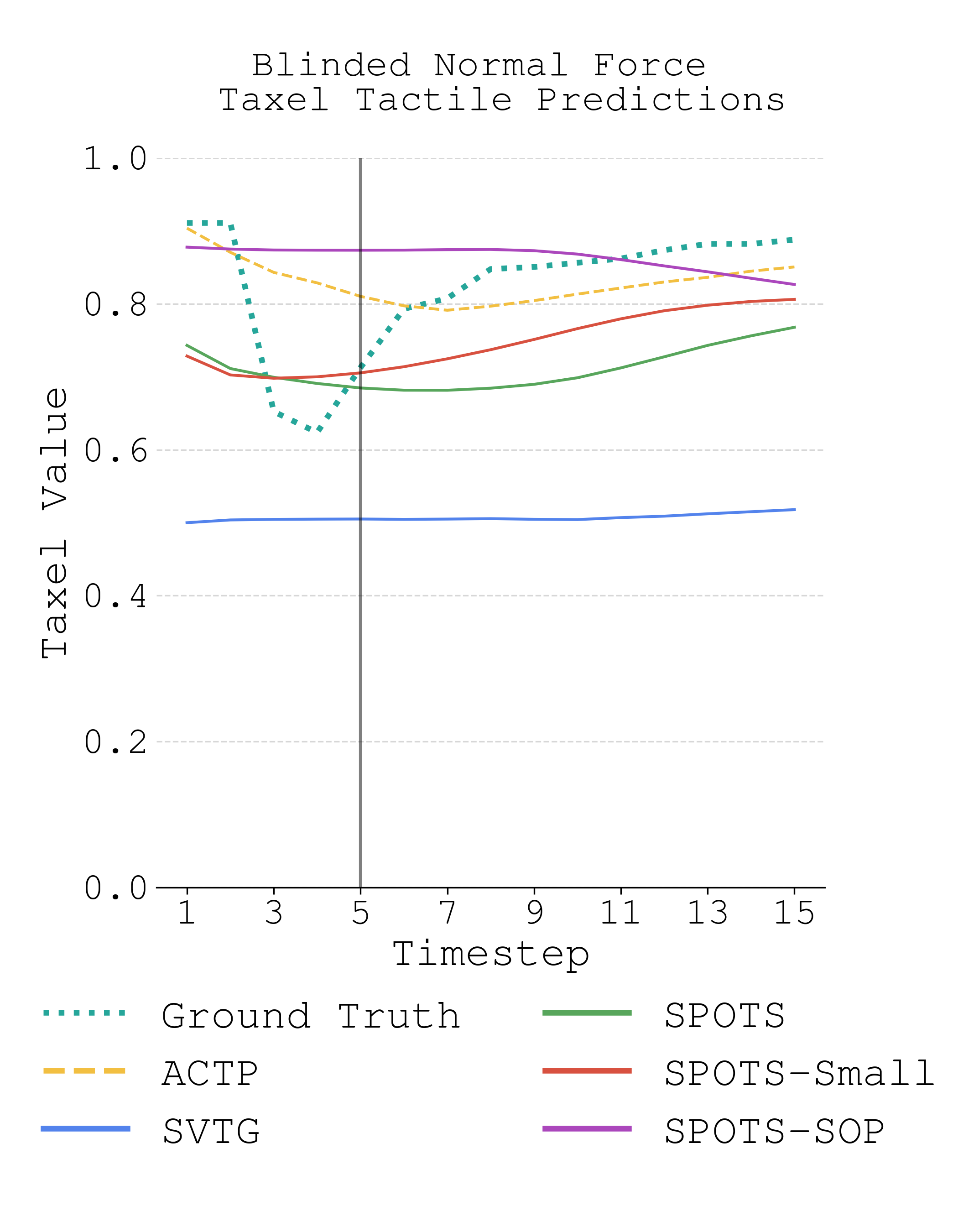}}}
    \caption{\textit{Visually Identical Edge Case Subset}: Tactile predictions with visually occluded scenes. Each graph shows a single Normal Force taxel from the center of the tactile sensor. The three prediction sequences shown are the same as in Fig. \ref{fig:Tactile_Prediction} (a, b \& c) for comparison. The models are trained to predict 5-time steps, indicated by the bold vertical black line, with an extended prediction horizon of 15-time steps shown. Without visual input, the multi-modal systems fail to predict spikes in tactile sensation, resulting in flat tactile data. This analysis indicates that multi-modal models rely heavily on visual features for accurate tactile predictions, outperforming the state-of-the-art uni-modal tactile prediction model, ACTP, under normal conditions.}
    \label{fig:Tactile_Prediction_VO}
\end{figure*}

\chdeleted[id=AE]{This subsection evaluates the tactile prediction capabilities of the multi-modal models. Table~\ref{tab:TacPredPerformance} reports quantitative performance measured by Mean Absolute Error (MAE) over the full prediction horizon as well as at selected time steps, while Fig.~\ref{fig:Tactile_Prediction} presents representative qualitative results.}
\chdeleted[id=AE]{The dual-pipeline SPOTS models consistently outperform both the tactile-only baseline (ACTP) and the single-pipeline multi-modal model (SVTG).}
\chdeleted[id=AE]{SVTG adopts an SVG-style architecture for predicting magnetic-based tactile signals, an approach that has previously been shown to be suboptimal for tactile prediction \cite{mandil2021tac}.}

\chdeleted[id=AE]{In contrast, the dual-pipeline SPOTS architecture allows each modality to be modelled using structures suited to its dynamics, enabling accurate prediction in both the tactile and visual domains.}  
\chdeleted[id=AE]{As a result, the single-pipeline SVTG model, despite performing reasonably for visual prediction, struggles to represent tactile dynamics due to its rigid shared architecture.} 
\chdeleted[id=AE]{Qualitative analysis in Fig.~\ref{fig:Tactile_Prediction} highlights the limitations of ACTP, which exhibits minimal variation in predicted force profiles over the prediction horizon.}  
\chdeleted[id=AE]{Consistent with its quantitative performance, SVTG produces tactile predictions that often deviate substantially from the ground truth force trajectories.}

\chreplaced[id=AE]{ The dual pipeline (SPOTS) models produce good estimates of changes in tactile sensation, predicting spikes at the right time steps and with roughly correct scale.} {In contrast, the SPOTS models accurately capture changes in tactile sensation, correctly predicting contact-induced force spikes at appropriate time steps and with approximately correct magnitude.} 
\chreplaced[id=AE]{Post spike the tactile predictions can linger at high/low values, for example in Fig \ref{fig:Tactile_Prediction} (a).} {Following contact events, predicted force values may occasionally linger at elevated or reduced levels (e.g., Fig.~\ref{fig:Tactile_Prediction}a), a limitation shared across all evaluated architectures.}

\chdeleted[id=AE] {Overall, differences between SPOTS variants are minor, indicating that the dual-pipeline design itself is the primary factor driving improved tactile prediction performance, while architectural refinements have a secondary effect. Overall, both quantitative and qualitative results demonstrate that SPOTS can generate realistic tactile predictions over extended horizons.}
\chdeleted[id=AE]{We next analyse the role of visual input in tactile prediction to determine whether visual information is only beneficial during training or whether it is actively leveraged during inference. To this end, we evaluate tactile prediction under visual occlusion by blinding the agent and replacing scene observations with a neutral grey image. Under visual occlusion, tactile prediction performance degrades to levels comparable with the tactile-only baseline, indicating that visual input is actively exploited during inference. Taken together with the anaesthetisation results, these findings demonstrate that multi-modal prediction relies on bidirectional cross-modal interaction, with tactile input refining visual predictions and visual input supporting tactile forecasting. This interdependence, established during training and maintained at inference time, enables each modality-specific pipeline to outperform its state-of-the-art unimodal counterpart, resulting in a more faithful representation of physical cause-and-effect relationships during robot interaction.}

%%%%%%% DISCUSSION %%%%%%%%
%%%%%%% DISCUSSION %%%%%%%%
%%%%%%% DISCUSSION %%%%%%%%
\subsection*{Discussion and Limitations}
\chreplaced[id=AE]{Both quantitative and qualitative analyses indicate that integrating tactile and visual input can improve physical robot interaction prediction, with the magnitude and nature of the improvement depending on the interaction regime and evaluation criterion~\cite{nazari2022proactive, nazari2023deep}.}{Both quantitative and qualitative results demonstrate that integrating tactile and visual input into physical robot interaction prediction models leads to more accurate scene and tactile predictions, which is useful in enhanced physical interaction~\cite{nazari2022proactive}.}

\chreplaced[id=AE]{Our results indicate that separating heterogeneous sensory modalities into modality-specific pipelines, while allowing limited cross-modal interaction, provides a more effective bias–variance trade-off than enforcing a shared latent representation.}{Quantitative scene analysis, which evaluates image accuracy on a pixel-by-pixel basis, reveals that the SPOTS multi-modal prediction architecture produces the most accurate predictions and generalizes better to new objects.}
\chdeleted[id=AE]{We attribute SPOTS' superior performance to its dual-pipeline system, which processes tactile data separately from visual data. This separation allows the visual pipeline to generate crisper predictions, whereas the single-pipeline SVTG model struggles by combining both modalities into one architecture, leading to significantly worse results.}

\chdeleted[id=AE]{Qualitative scene analysis, focusing on predicted object locations, indicates that all tactile-integrated architectures achieve more accurate predictions in visually identical scenes with objects of varying physical properties. Interestingly, SVTG, the worst-performing architecture in quantitative analysis, excels in qualitative analysis, suggesting a trade-off between realistic visual predictions and accurate physical interaction perception. SVTG's object location predictions are the most precise, although its predictions of objects' physical structure are the least visually realistic.}
\chdeleted[id=AE]{Tactile prediction analysis, both qualitative and quantitative, further underscores the advantages of the dual-pipeline SPOTS architecture over both the state-of-the-art tactile prediction architecture ACTP and the single-pipeline multi-modal SVTG model. While the qualitative performance of the single- and dual-pipeline architectures could not be distinguished in visual prediction tasks, they were clearly separated in tactile prediction tasks due to SVTG's poor ability to predict tactile sensations. The dual-pipeline architecture allows the best aspects of both modalities to be retained.}
\chadded[id=AE]{Taken together, these findings reinforce a general design principle for multi-modal prediction: forcing heterogeneous sensory modalities into a single shared latent space can compromise modality-specific fidelity, whereas maintaining separate prediction pipelines with limited cross-modal coupling provides a more effective bias--variance trade-off.}

\chreplaced[id=AE]{Overall, the optimal architecture depends on the downstream task and the relative importance of visual fidelity versus physical interaction understanding. For applications requiring both realistic visual predictions and accurate tactile forecasting, SPOTS is the preferred architecture. In contrast, for vision-dominant control tasks where tactile prediction is secondary, SVTG may be advantageous due to its strong localisation performance. Given its balanced performance across modalities and its architectural flexibility, we consider SPOTS a promising foundation for future multi-modal prediction research.}{Overall, the optimal architecture depends on the downstream application. For realistic visual predictions and accurate tactile predictions, SPOTS is the preferred architecture. However, for vision-based control tasks, SVTG may be more suitable due to its slightly better physical interaction perception, despite its poor tactile predictions. Given its accuracy in both modalities and its adaptability to different modalities, we consider the SPOTS architecture the most promising for future multi-modal research.}

\chreplaced[id=AE]{Regarding limitations, performance gains on the household cluster dataset were modest across all evaluated models. %Although SPOTS exhibited slightly improved generalisation to unseen objects, the absolute improvement in this visually unambiguous setting remained limited, as expected from the regime-based analysis in Section~6.1. 
As established in Section 6.1, the benefits of tactile–visual integration emerge most clearly under physical ambiguity rather than visually inferable interactions. These findings are consistent with the regime-based analysis in Section~6.1. Future work should explore scaling both model capacity and dataset size for the household cluster setting, which currently contains approximately 240{,}000 frames and may benefit from expansion toward larger-scale datasets (e.g., 1.5 million frames~\cite{ebert2017self}). Alternatively, reducing object diversity could enable more controlled experiments and faster iteration during model development.}
{Regarding limitations, the household cluster dataset yielded poor prediction results across all models tested. Although the SPOTS architecture showed marginally better generalisation to unseen objects at test time compared to the vision-only system, the improvement was minimal. Most conclusions about the benefits of tactile sensation in physical interaction perception were drawn from the second dataset. Future work should aim to develop more capable models for the household cluster dataset, potentially by increasing the dataset size, which currently consists of 240,000 frames but could benefit from expansion to match datasets as large as 1.5 million frames \cite{ebert2017self}. Reducing the number of objects in the dataset could also facilitate meaningful comparisons while allowing faster training for model development.}

\chreplaced[id=AE]{Parameter efficiency also plays a role in prediction performance. Notably, SPOTS-small achieves comparable or superior results to SPOTS despite its reduced parameter count, suggesting that architectural design is more influential than model size alone. In contrast, SVTG introduces a substantial increase in parameters due to single-pipeline modality fusion. Moreover, SPOTS incorporates a relatively lightweight tactile prediction module based on ACTP (approximately 14\% of total parameters). Future work could explore more parameter-efficient single-pipeline designs to disentangle architectural effects from model capacity. }{Parameter size also affects prediction performance. The SPOTS-small model outperformed SPOTS, suggesting that a smaller parameter count can be advantageous. The SVTG model has a larger parameter size due to the added modality, with hidden layers doubled from 256 to 512 in the frame prediction network. In contrast, the SPOTS system incorporates the ACTP prediction model for its tactile prediction pipeline, which is relatively small (accounting for approximately 14\% of the network parameters). A limitation of this study is that the large parameter size of the SVTG model may negatively impact its qualitative performance; future research could explore whether a smaller hidden layer size might yield better results.}

\chreplaced[id=AE]{Other future work will focus on identifying which tactile features are most informative for physical interaction perception, starting from low-resolution sensing. While this study employs recurrent architectures, extending the proposed multi-modal prediction framework to transformer-based models is a promising direction, particularly when larger-scale interaction datasets are available.}{In future work, we recommend exploring which features and attributes of tactile sensation are most useful for physical interaction perception, using a simple low-resolution sensor as a starting point. Future work could also explore the application of multi-modal prediction in physical robot interaction using \emph{Transformer} models. These models, known for their complexity and substantial data requirements, have the potential to outperform the smaller models used in this study when trained on larger datasets. We anticipate that the integration of cross-modal information in such advanced models will further enhance the performance of unimodal predictions, leveraging the strengths of each modality to achieve superior results as we discussed here.}

\chadded[id=AE]{Finally, Recent work has shown that tactile representations play a critical role in downstream manipulation tasks such as slip detection and grasp stability assessment \cite{elijah2025advancing}. Integrating such tactile reasoning with predictive world models represents a promising direction for future research, enabling more robust and anticipatory manipulation strategies.
}

%%%%%%%%%%%%%%%%%%%%%%%%%%%

\section{Conclusion}
\chreplaced[id=AE]{In conclusion, we presented a multi-modal prediction framework for physical robot interaction that jointly models visual and tactile dynamics. Through systematic comparison of multiple architectures, we demonstrated that tactile input can improve video prediction accuracy during physical interaction, while visual input can likewise enhance tactile prediction. Crucially, we showed that simultaneous prediction of tactile and visual signals yields the greatest improvements, indicating that active cross-modal interaction enables one modality to refine predictions in the other, particularly under physical ambiguity.}{ Our multi-modal approach, tested with various model architectures, demonstrates that incorporating tactile input enhances video prediction accuracy during robot pushing, while visual input similarly improves tactile prediction accuracy. Additionally, we showed that the simultaneous prediction of tactile and visual data has the greatest positive impact, indicating that multi-modal prediction models can leverage predictions from one modality to enhance performance and physical interaction perception in the other. }

\chreplaced[id=AE]{By utilising cross-modal connections between modality-specific prediction pipelines, the proposed dual-pipeline SPOTS architecture achieves improved
performance over strong tactile-only (ACTP) and vision-only (SVG) baselines, while preserving visual fidelity and tactile prediction accuracy.}{By utilising cross-modal connections between modality-specific prediction models, the dual-pipeline SPOTS architecture outperforms advanced tactile and video prediction models, i.e. ACTP and SVG, in their respective tactile and visual tasks.}
\chreplaced[id=AE]{While our study focuses on robot pushing as a representative interaction, the results suggest that multi-modal prediction is a promising direction for a wider range of physical interaction tasks, including object grasping and
manipulation, human--robot interaction, and tactile exploration. The proposed multi-modal approach also improves robustness, enabling one modality to partially compensate for missing or degraded information in the other.}{While our work introduces baseline approaches, we believe that the potential benefits of multi-modal prediction systems warrant further exploration across a range of physical robot interaction tasks, such as object grasping and manipulation, human-robot interaction, and tactile exploration. The multi-modal approach also offers increased robustness, allowing one modality to compensate for occlusions in the other. }

\chreplaced[id=AE]{Beyond vision and touch, incorporating additional sensory modalities such as audition may further enhance physical interaction perception and cause--effect understanding. Future work may also explore a broader range of tactile sensing technologies and attributes, including temperature, vibration, and texture. In addition, integrating multi-modal prediction with model predictive
control offers an exciting avenue for closing the perception--action loop
in physical robot interaction.}{We also see potential in introducing auditory input to the SPOTS architecture, which may further enhance an agent's physical interaction perception and cause-effect understanding, enabling more complex interactions. Future work may also involve exploring a wide range of tactile sensing devices and their attributes, such as temperature, vibration, and texture sensing. Additionally, the integration of vision and tactile model predictive control presents an exciting avenue for robotics research.}

\chadded[id=AE]{An important direction for future work is integrating multi-modal world models directly into policy learning or model-based control loops, enabling hybrid predictive-control architectures where predicted cross-modal signals inform action selection in real time.}

\chreplaced[id=AE]{More broadly, our results highlight the potential of unsupervised multi-modal prediction for learning physical interaction models without task-specific supervision, facilitating scalable and cost-effective model development. Overall, this work demonstrates that carefully designed multi-modal prediction architectures can improve physical interaction perception, particularly in physically ambiguous settings, and provides a foundation for future research in multi-modal robot learning.}{Our approach to prediction in physical robot interaction opens up a wide range of future research opportunities, and we believe that the unsupervised learning approach makes the development of models in this domain both straightforward and cost-effective, facilitating advancements in various robotic fields. Overall, our work highlights the potential of multi-modal prediction models for physical interaction perception and presents exciting opportunities for future research in this field.}

\section*{Acknowledgments}
This work was partially supported by the UK Centre for Doctoral Training in Agri-Food Robotics (AgriFoRwArdS) (Grant reference: EP/S023917/1). The authors thank Jon Flynn, Karoline Heiwolt, and Kiyanoush Nazari for their valuable discussions. We also acknowledge the support of Sheffield Robotics.

 % argument is your BibTeX string definitions and bibliography database(s)
\bibliographystyle{elsarticle-harv.bst}
\bibliography{references.bib}
%
% \section{Simple References}
% You can manually copy in the resultant .bbl file and set second argument of $\backslash${\tt{begin}} to the number of references
%  (used to reserve space for the reference number labels box).

% \newpage

\end{document}